\pdfoutput=1
\documentclass[11pt]{article}

\usepackage{acl}

\usepackage{times}
\usepackage{latexsym}

\usepackage[T1]{fontenc}
\usepackage[utf8]{inputenc}

\usepackage{microtype}

\usepackage{inconsolata}

\usepackage{hyperref}
\usepackage{url}

\usepackage{amsmath}
\usepackage{amsfonts}
\usepackage{multirow}
\usepackage{booktabs,arydshln}
\usepackage{graphicx}
\usepackage{algorithm}
\usepackage{algorithmic}
\usepackage{tabu}
\usepackage{xcolor}
\usepackage{amssymb}
\usepackage{subfigure}

\makeatletter
\def\adl@drawiv#1#2#3{%
        \hskip.5\tabcolsep
        \xleaders#3{#2.5\@tempdimb #1{1}#2.5\@tempdimb}%
                #2\z@ plus1fil minus1fil\relax
        \hskip.5\tabcolsep}
\newcommand{\cdashlinelr}[1]{%
  \noalign{\vskip\aboverulesep
           \global\let\@dashdrawstore\adl@draw
           \global\let\adl@draw\adl@drawiv}
  \cdashline{#1}
  \noalign{\global\let\adl@draw\@dashdrawstore
           \vskip\belowrulesep}}
\makeatother

\newcommand{\rom}[1]{\uppercase\expandafter{\romannumeral #1\relax}}

\newcommand{\specialcell}[2][c]{\begin{tabular}[#1]{@{}c@{}}#2\end{tabular}}

\title{Improving In-context Learning of Multilingual Generative Language Models with Cross-lingual Alignment}
\author{Chong Li, Shaonan Wang, Jiajun Zhang\footnotemark[1], Chengqing Zong \\
        State Key Laboratory of Multimodal Artificial Intelligence Systems, \\
        Institute of Automation, CAS, Beijing, China\\
        School of Artificial Intelligence, University of Chinese Academy of Sciences, Beijing, China\\
        lichong2021@ia.ac.cn,\\
        \{shaonan.wang, jjzhang, cqzong\}@nlpr.ia.ac.cn
        }

\begin{document}
\maketitle

\renewcommand{\thefootnote}{\fnsymbol{footnote}} %将脚注符号设置为fnsymbol类型，即特殊符号表示
\footnotetext[1]{Corresponding author.}

\renewcommand{\thefootnote}{\arabic{footnote}}

\begin{abstract}
Multilingual generative models obtain remarkable cross-lingual in-context learning capabilities through pre-training on large-scale corpora. 
However, they still exhibit a performance bias toward high-resource languages and learn isolated distributions of multilingual sentence representations, which may hinder knowledge transfer across languages. 
To bridge this gap, we propose a simple yet effective cross-lingual alignment framework exploiting pairs of translation sentences. 
It aligns the internal sentence representations across different languages via multilingual contrastive learning and aligns outputs by following cross-lingual instructions in the target language. 
Experimental results show that even with less than 0.1 {\textperthousand} of pre-training tokens, our alignment framework significantly boosts the cross-lingual abilities of generative language models and mitigates the performance gap. 
Further analyses reveal that it results in a better internal multilingual representation distribution of multilingual models. 
\footnote{Our code is available at \href{https://github.com/chongli17/CrossLingualAlignment}{https://github.com/chongli17/\\CrossLingualAlignment}}
\end{abstract}

\section{Introduction}

\begin{figure}[th]
\centering
% \begin{minipage}[t]{0.5\linewidth}
\subfigure[$\text{XGLM}_{\text{564M}}$]{\label{fig:tsne_xglm_enzh}\includegraphics [scale=0.28]{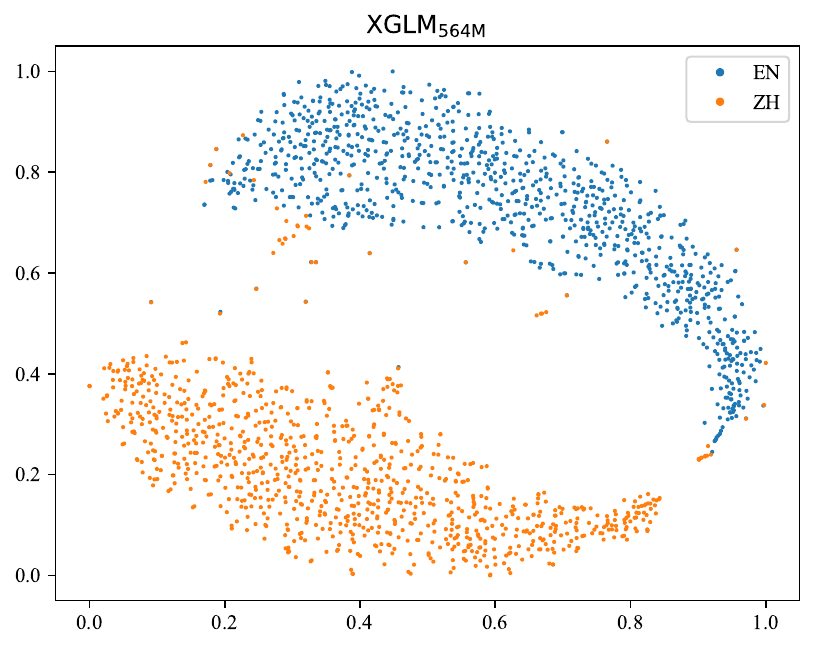}}
\subfigure[$\text{XGLM}_{\text{564M}}$ + AFP]{\label{fig:tsne_xglm_afp_enzh}\includegraphics [scale=0.28]{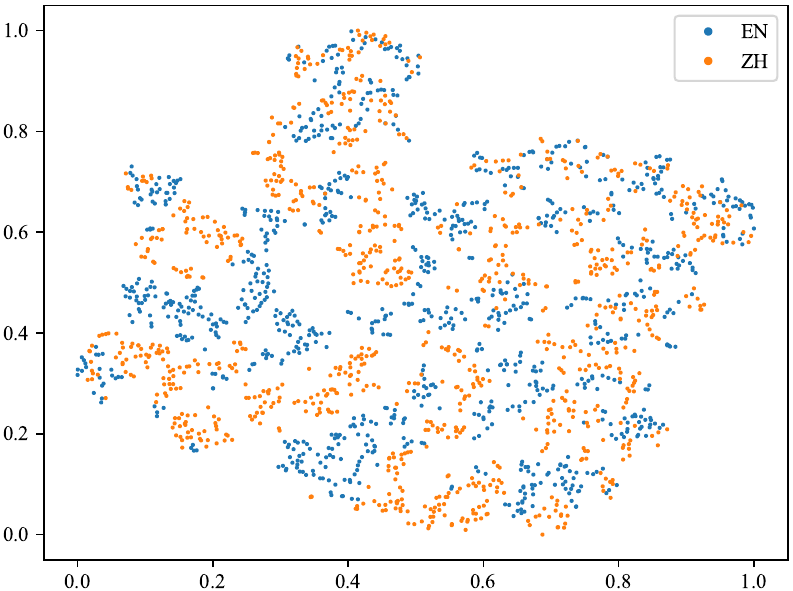}}
% \subfigure[]{\label{fig:tsne_xglm_enzh}\includegraphics [scale=0.25]{imgs/xglm-564m.en-zh.L1.pdf}}
% \subfigure[]{\label{fig:tsne_xglm_afp_enzh}\includegraphics [scale=0.25]{imgs/xglm-564m-afp.en-zh.L1.pdf}}
% \end{minipage}
% \begin{minipage}[t]{0.3\linewidth}
\subfigure[In-context Learning on XNLI]{\includegraphics [scale=0.28]{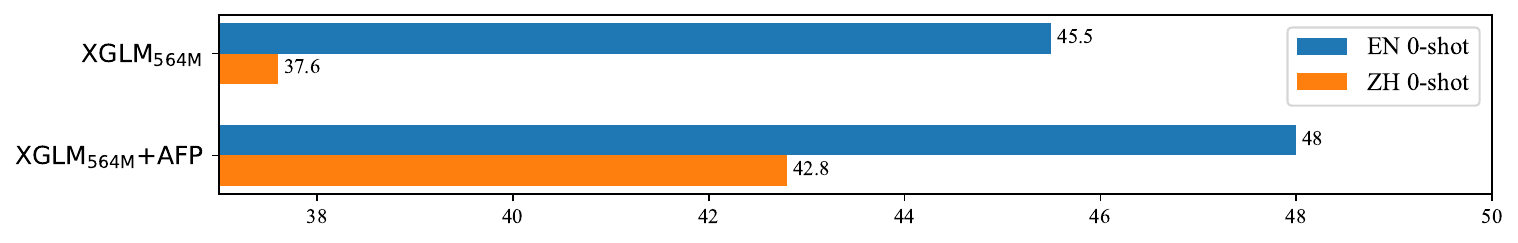}\label{fig:xglm_afp_onxnli_enzh}}
% \end{minipage}
\vspace{-3mm}
\caption{(a, b) Our method aligns the internal EN-ZH sentence representations of $\text{XGLM}_{\text{564M}}$, which are shown in t-SNE. (c) It also mitigates the performance gap on XNLI. }
\vspace{-3mm}
\end{figure}

Multilingual generative language models achieve impressive universality across many languages by pre-training on large-scale unsupervised multilingual corpora \citep{liu2020mbart, xue-etal-2021-mt5, lin-etal-2022-shot, scao2022bloom, soltan2022alexatm, openai2022chatgpt}. 
However, models still show a strong language bias toward high-resource languages \citep{asai2023buffet}, even the state-of-the-art multilingual generative models like GPT-4, exhibiting a 27.5\% relative performance gap between English and Telugu in MMLU \citep{openai2023gpt4}. 
This challenge partly arises from the significant linguistic resource imbalance among languages, which is hard to address solely through corpus scaling or balancing. 
Given such a model with language bias and the huge cost of re-training, how can we improve its cross-lingual capabilities and alleviate the language bias using limited data?

Previous work focused on scaling multilingual instructions \citep{muennighoff-etal-2023-crosslingual, zhu2023extrapolating}, ignoring internal alignment and knowledge transfer between languages in the multilingual generative model. 
Through visualizing the sentence representations in the multilingual generative model by mean pooling, we find that there is a distinct gap between the sentence representation distributions for different languages like Figure \ref{fig:tsne_xglm_enzh} (the multilingual ones are shown in Appendix \ref{appendix:multilingual_dist}). 
This is similar to learning representations for each language separately in the model, which is more challenging for multilingual models to transfer the knowledge learned from other languages. 
Thus, it is important to investigate whether the cross-lingual ability of models will be promoted by learning a better-aligned representation distribution. 

To address the above issues, we propose a cross-lingual alignment framework named \underline{A}lign a\underline{F}ter \underline{P}re-training (\textbf{AFP}), which aims to exploit translation pairs to narrow the gap between languages in the multilingual generation model. 
To be specific, our method can be divided into the following two modules: 1) \textbf{Multilingual Contrastive Learning (MCL)} on internal representations: we treat a translation sentence pair as the positive example for contrastive learning, and pull the sentence representations in two languages to be closer within the multilingual generated model. 
This module intends to reduce the differences between languages from the internal representations of the model. 
2) \textbf{Cross-lingual Instruction Following (CIF)} on the outputs: models must learn to answer in the target language given a prompt from the source language. 
It aims at enhancing semantic coherence and knowledge transfer across languages in the model. 

After extensive experiments, it can be found that AFP greatly improves the performance of multilingual generative models in cross-lingual natural language inference, multilingual reasoning, and other tasks using less than 1M parallel samples. 
The performance gap between languages is narrowed, e.g., the relative performance gap of $\text{XGLM}_{\text{564M}}$ reduces 6.53\% on XNLI between English and Chinese (Figure \ref{fig:xglm_afp_onxnli_enzh}). 
Our method also advances the performance on unseen languages for models, e.g., the Chinese performance of Llama, which is pre-trained on the corpus mainly in English \citep{touvron2023llama,touvron2023llama2}. 
Further analyses reveal that the representation gap has been mitigated as illustrated in Figure \ref{fig:tsne_xglm_afp_enzh} after training with AFP. 
In addition, experimental results show that the cross-lingual instruction following task is better than the multilingual instruction tuning task in promoting cross-lingual ability with the same parallel corpus. 

To sum up, our main contributions are as follows:
\begin{itemize}
    \item We propose a simple yet effective cross-lingual alignment framework, including the internal representation alignment (MCL) and external output alignment (CIF), to exploit the parallel corpus for multilingual generative models. 

    \item Experimental results demonstrate that our method greatly improves the cross-lingual ability of generative models, including multilingual ones and models pre-trained on English corpus, by using less than 1M samples.

    \item Further analyses reveal that AFP promotes the alignment and uniformity of internal multilingual representation distributions. Ablation study shows that using the internal representation alignment of AFP alone cannot boost multilingual generative models. 
    
    % We conduct qualitative and quantitive analyses on the internal multilingual representation distributions and find better alignment and uniformity after using AFP. 

    % \item Our framework can align the internal multilingual representation distributions and promote the knowledge transfer to the unseen languages of multilingual generative models. 
    
    % Further ablation study reveals that using multilingual contrastive learning alone cannot boost multilingual generative models. 
    % After alignment, models can also be applied with other methods to enhance performance further. 
    
\end{itemize}

\section{Related Work}
\subsection{Multilingual Generative Language Model}
Through unsupervised pre-training on the large-scale multilingual corpus, generative language models obtain impressive multilingual abilities, e.g., multilingual machine translation \citep{liu2020mbart, he2021synchronous, wang2022synchronous, lu-etal-2023-take}, cross-lingual natural language understanding \citep{xue-etal-2021-mt5} and cross-lingual in-context learning \citep{lin-etal-2022-shot, scao2022bloom, wei2023polylm, anil2023palm2}. 
Most of them extended the pre-training method developed for the monolingual corpus \citep{lewis-etal-2020-bart,raffel2020t5} and relied on a balanced sampling method across languages, while a significant performance gap between high-resource languages and low-represented languages persists in the pre-trained model \citep{asai2023buffet}. 
Different from the unsupervised pre-training on the multilingual corpus, this work attempts to alleviate the performance gap across languages by cross-lingual alignment using parallel samples. 

\subsection{Multilingual Instruction Tuning}
Large language models show better zero-shot multilingual performance and language generalization results after multilingual instruction tuning \citep{muennighoff-etal-2023-crosslingual, zhang2023bayling, zhu2023extrapolating, ranaldi2023empowering}. 
Our cross-lingual instruction following task requires the model to respond in the target language and is different from multilingual instruction tuning, where prompt and answer in the same language for each sample. 

\subsection{Contrastive Learning in Natural Langauge Processing}
Most of the work in NLP adopted contrastive learning to improve the representation of sentences in the language model \citep{reimers-gurevych-2019-sentence, pan-etal-2021-multilingual, gao-etal-2021-simcse, yang-etal-2021-xmoco, pan-et-al-2022-improved, ni-etal-2022-sentence, sherborne-etal-2023-optimal}. 
Specifically, contrastive learning is often applied to the sentence representations of encoder \citep{Cao2020Multilingual, fang2020cert, wu2020clear, pan-etal-2021-contrastive, chi-etal-2021-infoxlm, wei2021on}. 
However, it is less explored how to promote the representation of Transformer decoder models \citep{vaswani2017attention, zhao2023transformer}. 
%, which becomes more important as large language models often adopt the decoder structure. 
%, which may arise from our finding in the ablation study that using multilingual contrastive learning alone cannot boost multilingual generative models (Section \ref{sec:ablation}). 
In this work, we try to improve the internal multilingual representation of the decoder models by multilingual contrastive learning rather than the one of encoder \citep{wang-etal-2021-aligning, qin-etal-2022-gl}.

\begin{figure*}[t]
\begin{center}
%\framebox[4.0in]{$\;$}
\includegraphics[width=0.92\textwidth]{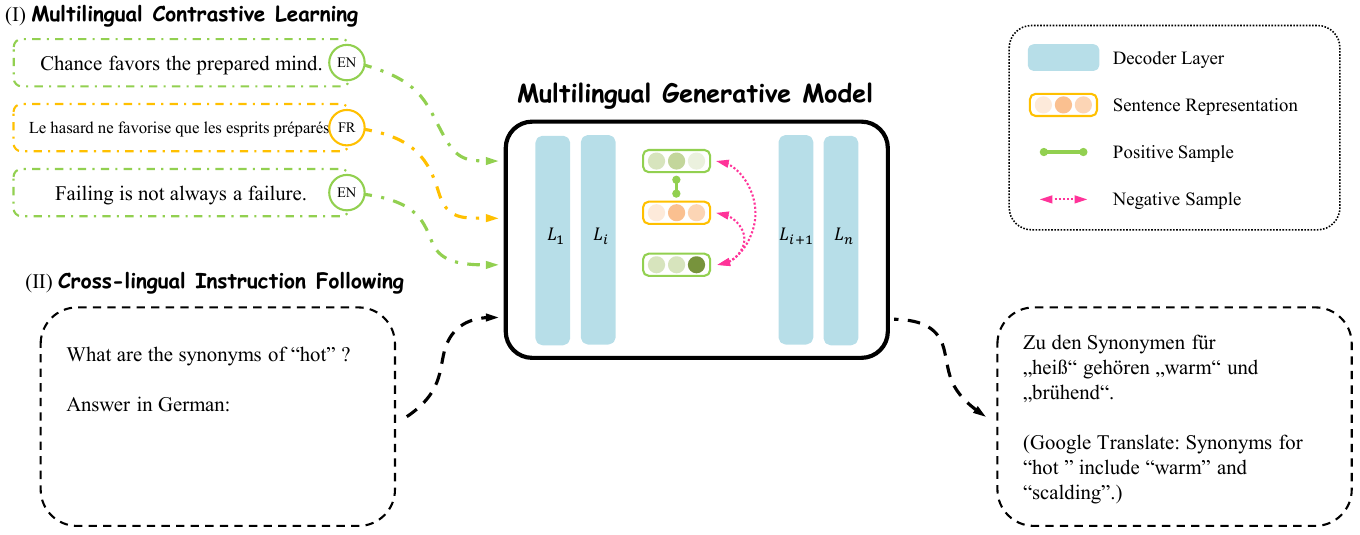}
\end{center}
\vspace{-4mm}
\caption{\label{fig:method}Illustration of how to align the internal representations and outputs of multilingual generative models with AFP. (\rom{1}) Given a translation parallel sample as the positive sample, multilingual contrastive learning pulls their representations together and pushes apart the ones from other samples.  (\rom{2}) Multilingual generative models are required to answer in the target language to align the outputs across languages.}
\vspace{-4mm}
\end{figure*}

\section{Method}
As shown in Figure \ref{fig:method}, our framework AFP contains the following two modules: 1) Multilingual contrastive learning (Section \ref{sec:MCL}), which aims to align the internal representations of models across different languages. 2) Cross-lingual instruction following (Section \ref{sec:CIF}), which requires models to align the outputs between different languages. 

\subsection{Multilingual Contrastive Learning}
\label{sec:MCL}
To align the internal multilingual representation of models, we exploit the contrastive learning method, which is generally found effective in aligning the representations from different modalities in multi-modal work \citep{radford21clip, xu-etal-2021-videoclip, liang2022mind}. 
Hence, translation pairs are regarded as positive instances with closely aligned semantics in multilingual contrastive learning, and we pull their internal representations closer. 
The other sentences in the same batch are taken as the negative samples for the translation pair. 

Formally, to align the $l$-th layer of model $f(\theta)$, the sentence representations $(h_i, h_i^{+})$ is calculated as follows:
\begin{equation}
h_i = g(f_{l}(s_i;\theta)),\ h_i^{+} = g(f_{l}(s_i^{+};\theta))
\end{equation}
where $f_{l}(\cdot)$ represents the output from the $l$-th layer, $g(\cdot)$ is the pooling method to obtain the sentence representation for decoder models, e.g., mean pooling or max pooling, and $(s_i, s_i^{+})$ is a parallel sample from $\mathcal{D}=\{(s_1, s_1^{+}), ..., (s_n, s_n^{+})\}$. 
We determine the specific layer to align according to the performance of models on the dev set and find that the first layer after embedding comes to better performance (please refer to Section \ref{sec:layer} for more details). 
Then, the training objective of Multilingual Contrastive Learning (MCL) is:
\begin{equation}
\small
\mathcal{L}_{\textit{MCL}}(\theta) = \underset{(s_i,s_i^{+})\thicksim \mathcal{D}}{\mathbb{E}} \left[ -{\text{log}\left( \frac{e^{\text{sim}(h_i,h_i^{+})/\tau}}{\sum_j{e^{\text{sim}(h_i,h_j)/\tau}}}\right)} \right]
\end{equation}
where $\text{sim}(\cdot)$ is used to determine the similarity between representations, which is cosine similarity in this work, $h_j$ is the sentence representation of $s_j$ in the mini-batch containing $(s_i, s_i^{+})$, and $\tau$ is a temperature hyper-parameter. 
% loss function and how to construct the positive samples and get the sentence embedding. 

\subsection{Cross-lingual Instruction Following}
\label{sec:CIF}
To further align the output of multilingual generative models, we introduce a method named Cross-lingual Instruction Following (CIF), which imposes models to respond in the target language given the source language as the context. 
It is more difficult than the multilingual instruction tuning task, which prompts and answers in the same language for each sample, and requires a better cross-lingual understanding and generation ability for multilingual generative models. 

Specifically, given a pair of context and response $(c_i^{a}, r_i^{a})$ from a Dataset $\mathcal{D}^{a}$ in the same language $a$, e.g., an English instruction tuning dataset like FLAN or Alpaca \citep{wei2022flan, wang-etal-2023-self-instruct, taori2023alpaca}, response $r_i^{a}$ is first translated into the target language $b$ by the translator $t^{a\to b}(\cdot)$. 
We append a prompt $p^{b}$ informing the target language $b$, e.g., ``Answer in German'' in Figure \ref{fig:method}, at the end of context to construct the training sample $\left(c_i^{a\to b}=c_i^{a}+p^{b}, r_i^{b}=t^{a\to b}(r_i^{a})\right)$ for CIF. 
Therefore, the loss function of CIF for the multilingual generative model $f(\theta)$ comes to:
\begin{equation}
\small
    \mathcal{L}_{\textit{CIF}}(\theta) = \underset{(c_i^{a}, r_i^{a})\thicksim \mathcal{D}^{a}}{\mathbb{E}} \left[ \sum_j {-\text{log}\left( \text{P}(r_{ij}^{b}|c_i^{a\to b},r_{i,<j}^{b};\theta)\right)} \right] 
\end{equation}
where the target language $b$ has the possibility $p_{\textit{src}} \in [0, 1]$ to be set the same as the source language $a$, which is a hyper-parameter and investigated in Section \ref{sec:p_src}. 
When the target language is always the source language of the context ($p_{\textit{src}} = 1$), it degenerates into the vanilla multilingual instruction tuning method. 

With the two modules of aligning methods mentioned before, Multilingual Contrastive Learning (MCL) and Cross-lingual Instruction Following (CIF), we obtain the following loss function of our alignment framework AFP:
\begin{equation}
\mathcal{L}_{\textit{AFP}}(\theta) = \mathcal{L}_{\textit{MCL}}(\theta) + \alpha \mathcal{L}_{\textit{CIF}}(\theta)
\label{eq:afp_loss}
\end{equation}
where $\alpha \in \mathbb{R}^{+}_0$ is a hyper-parameter to balance the two alignment methods. 

\section{Experiments}
\subsection{Experiments Settings}
\paragraph{Parallel Corpora}
To cover more parallel samples from different domains and languages, we adopt a multilingual instruction tuning dataset named Bactrian-X \citep{li2023bactrian}, which is translated into 52 languages from Alpaca \citep{taori2023alpaca} and Dolly \citep{DatabricksBlog2023dolly} by Google Translate, and a multilingual machine translation dataset, OPUS-100 \citep{zhang-etal-2020-improving}, to align the models evaluated. 
Only 100k parallel samples are selected from OPUS-100 in our experiments to match the amount of Bactrian-X, which contains 67k samples for each language. 
The number of tokens used is about 20M, which is nearly 0.05 {\textperthousand} of tokens used in the pre-training of BLOOM \citep{scao2022bloom}.  

\paragraph{Language models} 
We apply AFP on two multilingual generative model structures, XGLM \citep{lin-etal-2022-shot} and BLOOM \citep{scao2022bloom}, across three different parameter amounts. 
% They are both pre-trained on large-scale unsupervised multilingual corpora with a more balanced sampling method across languages. 
The models fine-tuning with multilingual instruction tuning, ``+MIT'' or BLOOMZ \citep{muennighoff-etal-2023-crosslingual}, are taken as the baseline. 
Llama \citep{touvron2023llama}, which is mainly pre-trained on English corpus, is also included for comprehensive evaluation. 
Training settings and hyperparameters are reported in Appendix \ref{appendix:hyper}. 

\paragraph{Multilingual Tasks} We evaluate the performance of models on the following benchmarks: 
\begin{itemize}
\itemindent=-5pt
\item \textbf{Natural Language Inference} We use XNLI \citep{conneau-etal-2018-xnli} in this task.
\item \textbf{Paraphrase Detection} PAWS-X \citep{yang-etal-2019-paws} is evaluated for this task. 
\item \textbf{Reasoning} We adopt XCOPA \citep{ponti-etal-2020-xcopa}, XStoryCloze \citep{lin-etal-2022-shot} and XWinograd \citep{tikhonov-ryabinin-2021-heads} in this task. 
\item \textbf{Machine Translation} For this task, we use FLORES-101 \citep{goyal-etal-2022-flores}. 
\end{itemize}
% We select a natural language inference dataset \citep{conneau-etal-2018-xnli}, a multilingual paraphrase detection dataset \citep{yang-etal-2019-paws}, a multilingual machine translation dataset \citep{goyal-etal-2022-flores}, two multilingual summarization dataset \citep{zhu-etal-2019-ncls, hasan-etal-2021-xlsum} and three commonsense reasoning datasets \citep{ponti-etal-2020-xcopa, lin-etal-2022-shot, tikhonov-2021-xwinogrande} for comprehensive evaluation.
The detailed descriptions and prompt formats for each task during evaluation are presented in Appendix \ref{appendix:prompt_templates}. 
We keep the same prompt formats across all multilingual generation models for a fair comparison. 

\begin{table*}[thp]

\renewcommand\arraystretch{1.1}

\centering
\scriptsize

\setlength{\tabcolsep}{0.9mm}

\vspace{-2mm}

 \begin{tabu}{l|c|c|c|c|c|c|c|c|c|c|c}
 
 \toprule[1.2pt]
  \multicolumn{1}{c}{\textbf{ }} & \multicolumn{2}{c}{\textbf{XNLI}} & \multicolumn{2}{c}{\textbf{PAWS-X}} & \multicolumn{2}{c}{\textbf{XCOPA}} & \multicolumn{2}{c}{\textbf{XStoryCloze}} & \multicolumn{2}{c}{\textbf{XWinograd}} & \\
  \cmidrule(r){2-3}  \cmidrule(r){4-5} \cmidrule(r){6-7} \cmidrule(r){8-9} \cmidrule(r){10-11} \noalign{\smallskip}
\multicolumn{1}{c}{\textbf{Model}}& \multicolumn{1}{c}{\textbf{EN-0/5}}&\multicolumn{1}{c}{\textbf{ZH-0/5}}& \multicolumn{1}{c}{\textbf{EN-0/5}}&\multicolumn{1}{c}{\textbf{ZH-0/5}}& \multicolumn{1}{c}{\textbf{EN-0/5}}&\multicolumn{1}{c}{\textbf{ZH-0/5}}& \multicolumn{1}{c}{\textbf{EN-0/5}}&\multicolumn{1}{c}{\textbf{ZH-0/5}} & \multicolumn{1}{c}{\textbf{EN-0/5}}&\multicolumn{1}{c}{\textbf{ZH-0/5}}& \multicolumn{1}{c}{\textbf{Avg}} \\

   \midrule[0.8pt]
   $\text{GPT-3}_{\text{6.7B}}$                  & $55.3/52.8$                      & $42.4/45.9$                       & $60.6/59.7$                   & $53.2/54.1$                   & $73.6/74.5$                   & $55.0/57.7$                   & $73.6/74.5$   & $55.9/54.5$   & $64.6/68.1$   & $71.5/72.2$ & $61.0$\\
   
  \midrule[0.8pt]
 
  $\text{XGLM}_{\text{564M}}$                  & $45.5/41.2$                      & $37.6/35.6$                       & $50.4/46.6$                   & $50.9/47.8$                   & $56.4/59.6$                   & $52.8/52.2$                   & $59.6/60.8$   & $54.3/52.9$   & $54.8/56.7$   & $67.1/66.9$ & $52.5$\\
   \ \ \ \ \ \ \ \ \ +MIT            & $46.6/43.9$                      & $37.5/41.6$                       & $53.5/53.1$                   & $52.3/51.0$                   & $57.6/61.0$                   & $57.2/55.4$                   & $61.1/61.3$   & $54.5/54.5$   & $\textbf{55.6}/57.7$   & $66.7/65.3$ & $54.4$\\
   
 \ \ \ \ \ \ \ \ \ +AFP            & $\textbf{48.1}/\textbf{46.5}$    & $\textbf{41.6}/\textbf{42.5}$     & $\textbf{54.2}/\textbf{53.8}$ & $\textbf{53.2}/\textbf{52.8}$ & $\textbf{62.0}/\textbf{62.2}$ & $\textbf{59.0}/\textbf{58.8}$ & $\textbf{62.2}/\textbf{62.5}$   & $\textbf{56.3}/\textbf{56.1}$   & $\textbf{55.6}/\textbf{59.0}$   & $\textbf{67.5}/\textbf{67.3}$ & $\textbf{56.1}$\\
    \cdashlinelr{1-12}

 % $\text{XGLM}_{\text{1.7B}}$                  & $47.6/42.9$                      & $39.0/39.3$                       & $51.0/52.4$                   & $51.1/48.8$                   & $62.4/64.0$                   & $56.4/57.8$                   & $62.7/64.5$   & $56.3/56.3$   & $57.2/60.0$   & $68.3/68.5$& $55.3$\\
  
 % \ \ \ \ \ \ \ \ \ +AFP            & $\textbf{48.3}/\textbf{46.9}$    & $\textbf{43.1}/\textbf{43.7}$     & $\textbf{55.2}/\textbf{55.7}$ & $\textbf{54.5}/\textbf{53.8}$ & $\textbf{66.8}/\textbf{68.8}$ & $\textbf{61.4}/\textbf{61.6}$ & $\textbf{65.5}/\textbf{66.7}$   & $\textbf{59.3}/\textbf{59.6}$   & $\textbf{58.8}/\textbf{64.7}$   & $\textbf{68.7}/\textbf{69.6}$& $\textbf{58.6}$\\
 %    \cdashlinelr{1-12}
 
  $\text{XGLM}_{\text{7.5B}}$                  & $54.1/49.9$                      & $45.4/44.2$                       & $58.9/56.3$                   & $52.9/55.8$                   & $69.4/74.6$                   & $62.4/63.2$                   & $69.2/73.7$   & $59.5/59.2$   & $62.8/66.4$   & $73.8/73.2$& $61.2$\\
  \ \ \ \ \ \ \ \ \ +MIT            & $54.3/54.1$                      & $47.8/44.8$                       & $63.1/57.3$                   & $54.4/55.0$    & $69.4/75.0$                   & $63.2/64.6$      & $71.1/74.3$   & $60.1/61.7$   & $64.5/67.5$   & $74.4/73.4$& $62.5$\\
  
 \ \ \ \ \ \ \ \ \ +AFP            & $\textbf{55.0}/\textbf{54.7}$    & $\textbf{48.0}/\textbf{48.8}$       & $\textbf{64.8}/\textbf{61.2}$ & $\textbf{57.8}/\textbf{56.4}$  & $\textbf{72.2}/\textbf{75.6}$ & $\textbf{64.4}/\textbf{66.8}$  & $\textbf{72.0}/\textbf{74.7}$   & $\textbf{62.7}/\textbf{63.4}$   & $\textbf{65.2}/\textbf{68.2}$   & $\textbf{75.8}/\textbf{74.0}$& $\textbf{64.1}$\\
 
 \specialrule{0em}{0pt}{0pt}

  \midrule[0.8pt]
$\text{BLOOMZ}_{\text{560M}}$                  & $43.8/44.5$                      & $41.5/40.7$                      & $52.4/51.2$                   & $54.1/52.9$                   & $54.8/57.2$                   & $52.0/52.8$                   & $\textbf{61.2}/\textbf{61.7}$   & $56.4/55.0$   & $54.8/55.4$   & $62.3/65.1$& $53.5$\\
$\text{BLOOM}_{\text{560M}}$                  & $44.4/40.4$                      & $41.1/40.3$                       & $50.5/52.3$                   & $49.0/49.4$                   & $53.0/57.4$                   & $49.8/54.0$                   & $55.2/58.2$   & $57.9/53.2$   & $54.3/55.6$   & $63.9/64.9$& $52.2$\\
  
 \ \ \ \ \ \ \ \ \ +AFP           & $\textbf{50.7}/\textbf{46.4}$    & $\textbf{47.5}/\textbf{44.8}$     & $\textbf{58.2}/\textbf{57.5}$ & $\textbf{54.9}/\textbf{54.8}$ & $\textbf{57.8}/\textbf{58.4}$ & $\textbf{52.6}/\textbf{55.4}$ & $57.0/59.0$   & $\textbf{59.7}/\textbf{58.3}$   & $\textbf{56.3}/\textbf{57.2}$   & $\textbf{64.7}/\textbf{65.2}$& $\textbf{55.8}$\\
    \cdashlinelr{1-12}
 
$\text{BLOOMZ}_{\text{1.7B}}$                  & $50.3/51.2$                      & $48.0/46.2$                       & $57.1/53.4$                   & $54.4/52.3$                   & $58.0/58.0$                   & $55.2/56.8$                   & $66.4/68.9$   & $59.8/62.3$   & $59.0/61.6$   & $66.1/67.7$& $57.6$\\
$\text{BLOOM}_{\text{1.7B}}$                  & $50.4/44.4$                      & $47.6/46.1$                       & $47.7/52.1$                   & $52.9/51.1$                   & $55.8/58.2$                   & $52.4/54.6$                   & $64.2/67.3$   & $60.1/60.6$   & $56.1/59.3$   & $67.9/65.9$& $55.7$\\

 \ \ \ \ \ \ \ \ \ +AFP           & $\textbf{52.9}/\textbf{51.3}$    & $\textbf{49.8}/\textbf{48.8}$     & $\textbf{61.0}/\textbf{58.0}$ & $\textbf{56.9}/\textbf{56.0}$ & $\textbf{60.8}/\textbf{61.6}$ & $\textbf{55.4}/\textbf{58.2}$ & $\textbf{66.4}/\textbf{69.0}$   & $\textbf{63.3}/\textbf{63.3}$   & $\textbf{59.3}/\textbf{60.7}$   & $\textbf{68.3}/\textbf{66.1}$& $\textbf{59.4}$\\
    \cdashlinelr{1-12}
 
 $\text{BLOOMZ}_{\text{7.1B}}$                  & $51.1/52.0$                      & $49.7/48.0$                      & $63.6/62.2$                & $56.9/56.1$                   & $61.2/62.4$                   & $57.6/59.8$                   & $\textbf{73.7}/\textbf{76.9}$   & $62.1/63.9$            & $\textbf{64.1}/\textbf{66.9}$   & $66.1/68.5$& $61.1$\\
 $\text{BLOOM}_{\text{7.1B}}$                  & $54.0/48.7$                      & $48.1/47.5$                       & $59.9/60.4$                         & $53.2/51.4$                   & $58.0/58.8$                   & $54.0/54.8$                   & $70.4/73.5$                     & $64.3/64.8$   & $60.6/63.8$   & $71.4/67.7$& $59.3$\\

 \ \ \ \ \ \ \ \ \ +AFP           & $\textbf{55.8}/\textbf{54.3}$    & $\textbf{50.2}/\textbf{50.4}$     & $\textbf{66.5}/\textbf{64.5}$ & $\textbf{58.7}/\textbf{56.8}$ & $\textbf{62.0}/\textbf{62.8}$ & $\textbf{58.2}/\textbf{61.0}$ & $72.9/75.6$   & $\textbf{68.0}/\textbf{68.6}$   & $62.9/66.2$   & $\textbf{73.0}/\textbf{70.8}$& $\textbf{63.0}$\\
 
 \specialrule{0em}{0pt}{0pt}

  \midrule[0.8pt]

 $\text{Bactrian-X}_{\text{7B}}$                  & $53.0/53.3$                      & $44.6/44.1$                      & $68.7/63.4$                & $56.7/53.6$                   & $76.8/85.8$                   & $54.4/55.2$                   & $79.5/83.3$   & $55.9/57.0$            & $75.0/80.6$   & $66.3/66.1$ & $63.7$\\
 
 $\text{ZH-Alpaca}_{\text{7B}}^{\ddagger}$                  & $51.7/52.9$                      & $47.2/46.2$                       & $67.6/62.8$                         & $57.2/54.8$                   & $73.2/83.8$                   & $\textbf{57.6}/\textbf{60.8}$                   & $76.6/79.3$                     & $\textbf{57.4}/\textbf{58.3}$   & $71.4/74.8$   & $\textbf{67.9}/\textbf{68.5}$& $63.0$\\
 $\text{Llama}_{\text{7B}}$                  & $54.5/49.0$                      & $45.9/44.9$                       & $67.8/64.2$                         & $55.4/53.1$                   & $74.6/84.2$                   & $55.8/57.4$                   & $77.0/80.7$                     & $55.0/55.5$   & $72.3/79.4$   & $66.1/65.5$& $62.9$\\

 \ \ \ \ \ \ \ \ \ +AFP           & $\textbf{55.9}/\textbf{54.1}$    & $\textbf{47.6}/\textbf{48.4}$     & $\textbf{70.0}/\textbf{64.3}$ & $\textbf{58.6}/\textbf{56.1}$ & $\textbf{78.4}/\textbf{86.8}$ & $57.2/60.0$ & $\textbf{79.9}/\textbf{84.0}$   & $56.8/57.6$   & $\textbf{76.4}/\textbf{83.0}$   & $66.7/67.7$& $\textbf{65.5}$\\

\bottomrule[1.2pt]
\end{tabu}

\vspace{-2mm}
\caption{\label{tab:zh_en_bilingual} In-context learning results of models across different parameter scales on 5 datasets. The Average improvement is 3.31\%, where 4.28\% on the first two tasks and 2.67\% on reasoning tasks. ${}^{\ddagger}$ uses an additional 20GB Chinese corpus for pre-training. For a fair comparison, all results are obtained from the same in-context learning template illustrated in Appendix \ref{appendix:prompt_templates}. 
}
\vspace{-2mm}

\end{table*}
\begin{table*}[ht]

\renewcommand\arraystretch{1.1}

\centering
\scriptsize

\setlength{\tabcolsep}{3.2mm}

 \begin{tabu}{l|cccc|cccc|cccc}
 
 \toprule[1.2pt]
  \multicolumn{1}{c}{\textbf{ }} & \multicolumn{4}{c}{\textbf{EN$\to$ZH}} & \multicolumn{4}{c}{\textbf{ZH$\to$EN}} & \multicolumn{4}{c}{\textbf{Avg}} \\
  \cmidrule(r){2-5} \cmidrule(r){6-9} \cmidrule(r){10-13}  \noalign{\smallskip}
\multicolumn{1}{c}{\textbf{Model}}& 0 & 1 & 5 & \multicolumn{1}{c}{10} & 0 & 1 & 5 & \multicolumn{1}{c}{10} & 0 & 1 & 5 &10 \\

   \midrule[0.8pt]
 
  $\text{XGLM}_{\text{564M}}$      & $25.3$&$31.6$&$62.2$&$63.3$&$26.7$&$67.4$&$69.8$&$70.8$&$26.0$&$49.5$&$66.0$&$67.1$\\

 \ \ \ \ \ \ \ \ \ +AFP            & $\textbf{52.7}$&$\textbf{62.8}$&$\textbf{65.4}$&$\textbf{67.6}$&$\textbf{65.9}$&$\textbf{70.9}$&$\textbf{71.8}$&$\textbf{72.3}$&$\textbf{59.3}$&$\textbf{66.9}$&$\textbf{68.6}$&$\textbf{69.9}$ \\

    \cdashlinelr{1-13}

 $\text{XGLM}_{\text{7.5B}}$       &$28.1$&$79.3$&$79.8$&$80.1$&$29.1$&$81.6$&$81.8$&$82.2$&$28.6$&$80.4$&$80.8$&$81.2$ \\

 \ \ \ \ \ \ \ \ \ +AFP            &$\textbf{57.4}$&$\textbf{80.4}$&$\textbf{80.8}$&$\textbf{81.0}$&$\textbf{68.8}$&$\textbf{81.7}$&$\textbf{81.8}$&$\textbf{82.4}$&$\textbf{63.1}$&$\textbf{81.1}$&$\textbf{81.3}$&$\textbf{81.7}$\\

 %  \midrule[0.8pt]

 %   $\text{BLOOM}_{\text{560M}}$             & $0.1$&$1.5$&$2.8$&$3.2$                                       & $0.2$&$1.8$&$3.3$&$3.5$                                        & $0.2$&$1.7$&$3.1$&$3.2$    \\
  
 % \ \ \ \ \ \ \ \ \ +AFP            & $\textbf{1.5}$&$\textbf{3.1}$&$\textbf{4.0}$&$\textbf{4.5}$   & $\textbf{2.2}$&$\textbf{3.6}$&$\textbf{4.3}$&$\textbf{4.9}$    & $\textbf{1.9}$&$\textbf{3.4}$&$\textbf{4.2}$&$\textbf{4.7}$    \\
 
 %    \cdashlinelr{1-13}

 % $\text{BLOOM}_{\text{7.1B}}$               & $2.0$&$8.1$&$8.2$&$9.7$                                       & $3.0$&$10.6$&$12.0$&$12.4$                                     & $2.5$&$9.3$&$10.2$&$11.1$   \\
  
 % \ \ \ \ \ \ \ \ \ +AFP            & $\textbf{4.5}$&$\textbf{9.8}$&$\textbf{10.2}$&$\textbf{10.5}$ & $\textbf{5.1}$&$\textbf{11.0}$&$\textbf{12.4}$&$\textbf{12.7}$ & $\textbf{4.8}$&$\textbf{10.4}$&$\textbf{11.3}$&$\textbf{11.6}$  \\
 
 % \specialrule{0em}{0pt}{0pt}

\bottomrule[1.2pt]
\end{tabu}

\vspace{-2mm}
\caption{\label{tab:2lang_gen} Translation results of COMET \citep{rei-etal-2020-comet} on FLORES-101 devtest set. 
}

\vspace{-4mm}

\end{table*}

\subsection{Bilingual Results and Analyses}
\label{sec:bi_res}
To make a comprehensive analysis of the influence on performance and representations in models, we first conduct bilingual alignment experiments in English and Chinese. 
Then we extend to the condition of multilingual alignment (Section \ref{sec:mu_res}).

Table \ref{tab:zh_en_bilingual} shows the experimental alignment results on EN-ZH parallel samples. 
These generative models, including three architectures with different amounts of parameters, are consistently improved by our method. 
The average improvement is up to 3.31\% using only 167k parallel samples, and the models with 7B parameters surpass the GPT-3 with comparable parameters after alignment. 
Specifically, models improve 4.28\% on the first two natural language understanding tasks (XNLI and PAWS-X), and 2.67\% on the other three reasoning tasks. 
After alignment using AFP, BLOOM shows a better performance than the BLOOMZ model with the same amount of parameters, which is fine-tuned on 78M multilingual instructions \citep{muennighoff-etal-2023-crosslingual}. 

It is interesting to find that the model Llama pre-trained on mainly English corpus, also obtains improvement after bilingual alignment using AFP. 
The performance on the unseen language Chinese is even comparable with the one pre-training on an additional 20GB Chinese corpus \citep{cui2023chinesealpaca}. 
This result further proves the effectiveness of our method. 
We assume that this performance gain may benefit from better-aligned multilingual representations in models, which promotes the transfer of knowledge learned in the English corpus. 

% en-zh of xglm(564m, 1.7b, 7b) and bloom (560m, 1.7b, 7b) on xnli, paws-x (nlu), xcopa, xstorycloze, xwinograd (reasoning)

\begin{figure}[t]
\flushleft
\subfigure[\label{fig:tsne_bloom_enzh}$\text{BLOOM}_{\text{560M}}$]{\includegraphics [scale=0.28]{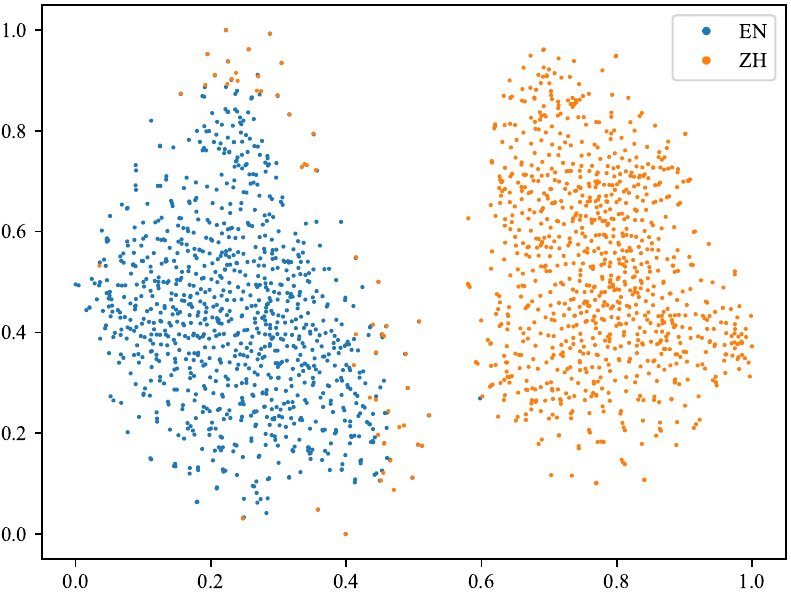}\vspace{-2mm}}
\subfigure[\label{fig:tsne_bloom_afp_enzh}$\text{BLOOM}_{\text{560M}}$+AFP]{\includegraphics [scale=0.28]{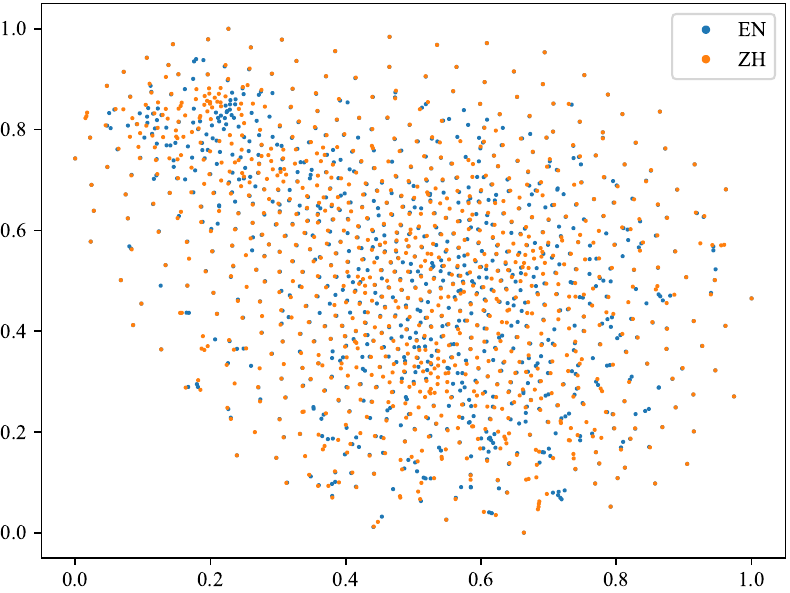}\vspace{-2mm}}
% \subfigure[${\mathcal{L}}_{\text{uniform}}\text{-}{\mathcal{L}}_{\text{align}}$]{\includegraphics [scale=0.31]{imgs/align_afp.pdf}\label{fig:xglm_align_afp_enzh}\vspace{-2mm}}
\vspace{-2mm}
\caption{(a, b) The t-SNE visualization of the original and aligned internal EN-ZH sentence representations of $\text{BLOOM}_{\text{560M}}$}
\vspace{-4mm}
\end{figure}

In addition to cross-lingual understanding and reasoning abilities, the multilingual generation ability of models has been improved. 
The bilingual translation results of XGLM models are reported in Table \ref{tab:2lang_gen}. 
Models not only obtain a better cross-lingual generation ability, but also show a more balanced generation performance than the vanilla ones between both directions. 
It is interesting to find that the average performance of models in the zero-shot condition improves from 27.3 to 61.2 COMET on average, which may come from the response in the target language format used in cross-lingual instruction following is similar to the one in the machine translation task. 

\subsubsection{AFP Brings Better Bilingual Representations}

\paragraph{Visualization of sentence representations.} Given 1k EN-ZH translation parallel samples, we visualize the sentence representations of $\text{XGLM}_{\text{564M}}$ and $\text{BLOOM}_{\text{560M}}$, which are obtained by the mean pooling method using the representation for each token in one sentence. 
In the vanilla models, there is a distinct separation between sentence representations from different languages (Figure \ref{fig:tsne_xglm_enzh} and \ref{fig:tsne_bloom_enzh}). 
However, the ones using AFP come to be more aligned between languages and uniform (Figure \ref{fig:tsne_xglm_afp_enzh} and \ref{fig:tsne_bloom_afp_enzh}), which means our method promotes the representation of the model to be better-aligned from a qualitative point of view. 
% The sentence representations of vanilla models across EN and ZH languages 
% The distribution of multilingual representation before and after alignment.

\paragraph{Alignment and uniformity.} The distribution of multilingual representations is quantified by the two metrics, \textbf{alignment} and \textbf{uniformity} proposed by \citet{wang202oalignment}, for further analysis. 
Specifically, the alignment score measures the expected distance between the representations of positive samples, which are translation parallel samples for multilingual generative models, and is calculated as follows:
\begin{equation}
    \ell_{\text{align}} \overset{\vartriangle}{=} \underset{(x,x^{+})\thicksim \mathcal{D}_{pos}}{\mathbb{E}} \left\lVert f(x) - f(x^{+}) \right\lVert^{2}
\end{equation}
where $\mathcal{D}_{pos}$ is the distribution of positive samples. 

\begin{figure}[t]
\includegraphics[width=0.42\textwidth]{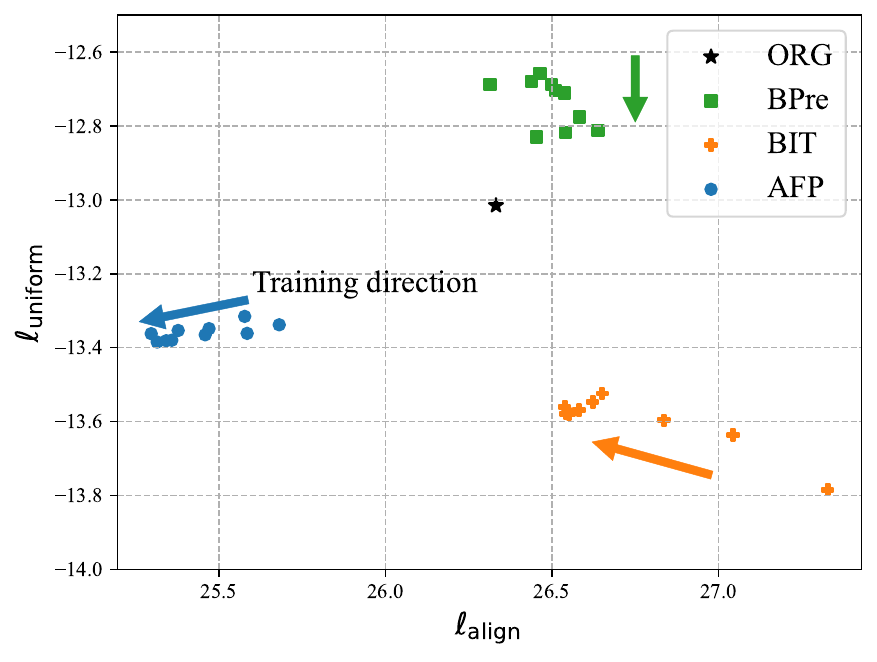}\vspace{-2mm}
\vspace{-2mm}
\caption{\label{fig:xglm_align_afp_enzh}The deviation of $\ell_{\text{uniform}}\text{-}\ell_{\text{align}}$ for $\text{XGLM}_{\text{564M}}$ during training process with different multilingual training methods. The smaller these two metrics are, the better representations models learn. ``BPre'' and ``BIT'' denote bilingual pre-training and bilingual instruction tuning, respectively.}
\vspace{-2mm}
\end{figure}

In contrast, uniformity reflects the degree of uniformly distributed for representations: 
\begin{equation}
    \ell_{\text{uniform}} \overset{\vartriangle}{=} \text{log} \underset{x,y \overset{i.i.d.}{\thicksim} \mathcal{D}}{\mathbb{E}} e^{-2\left\lVert f(x) - f(y) \right\lVert^{2}} 
\end{equation}
where $x$ and $y$ are randomly sampled from the distribution $\mathcal{D}$. 
Therefore, the smaller $\ell_{\text{align}}$ and $\ell_{\text{uniform}}$ are, the better representations models learn.

Figure \ref{fig:xglm_align_afp_enzh} illustrates the deviation of $\ell_{\text{align}}$ and $\ell_{\text{uniform}}$ for $\text{XGLM}_{\text{564M}}$ using different training methods on the same training data. 
The initial 5000 steps are visualized, with one point for every 500 steps. 
We can find that the metrics are both decreasing using AFP, while the bilingual pre-training only improves the uniformity of representations. 
The results further prove that our method improves the multilingual representation distributions within the multilingual generative models.

\subsubsection{Multilingual Contrastive Learning on Bottom Layer Performs Better}
\label{sec:layer}
\begin{figure}[t]
% \flushleft
\centering
\subfigure[]{\includegraphics [scale=0.25]{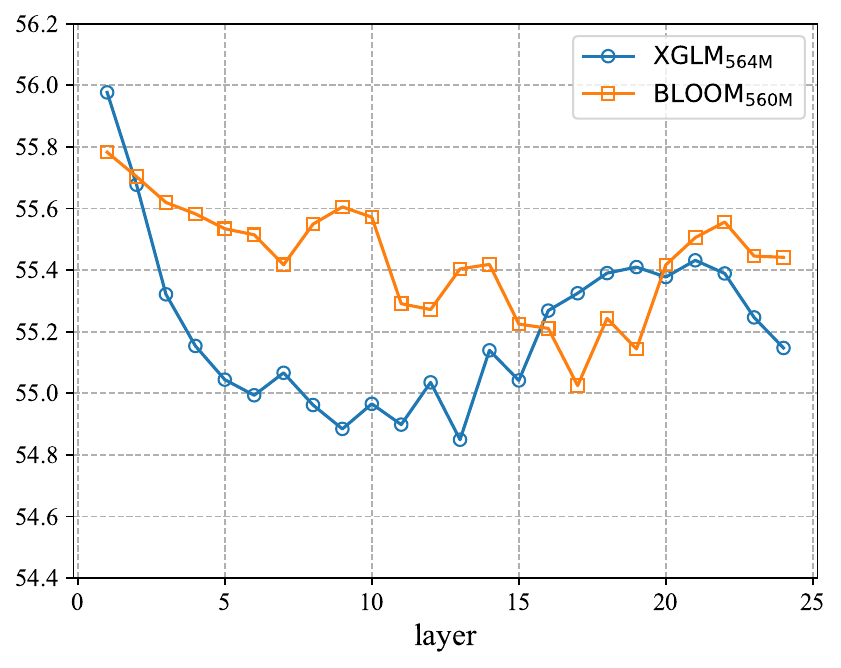}\label{fig:mcl_layer}}
\subfigure[]{\includegraphics [scale=0.25]{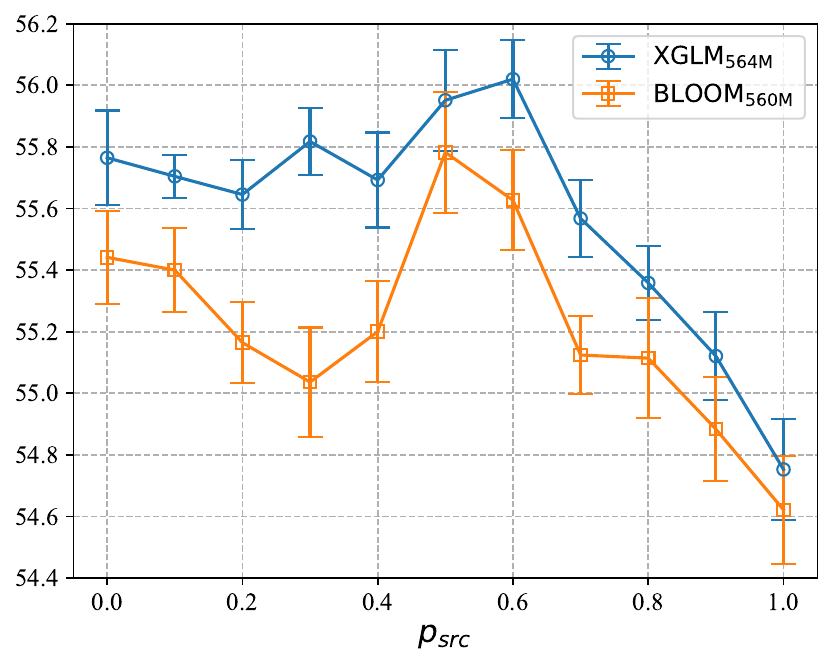}\label{fig:cif_p_cross}}
\vspace{-2mm}
\caption{Effects of the target layer of MCL (a) and the $p_{\textit{src}}$ of CIF (b) on 5 EN-ZH datasets. }
\vspace{-4mm}
\end{figure}

Figure \ref{fig:mcl_layer} presents the impact of different layers applied by contrastive learning on the 5 cross-lingual datasets (XNLI, PAWS-X, XCOPA, XStoryCloze, and XWinograd). 
The average performance of models shows a trend of decreasing first and then increasing, which changes at the 10th layer for $\text{XGLM}_{\text{564M}}$ or the 17th layer for $\text{BLOOM}_{\text{560M}}$. 
And the first transformer layer is better for both models when using multilingual contrastive learning. 
As a result, multilingual contrastive learning is applied to the first layer after the embedding layer by default. 

% \begin{figure}[th]
% \includegraphics [width=0.4\textwidth]{imgs/layers.pdf}
% \vspace{-2mm}
% \caption{\label{fig:mcl_layer}Effects of the target layer of MCL}
% \vspace{-2mm}
% \end{figure}

% \begin{figure}[th]
% \includegraphics [width=0.4\textwidth]{imgs/p_cross.pdf}
% \vspace{-2mm}
% \caption{\label{fig:cit_p_cross}$p_{\textit{src}}$ of CIT on 5 EN-ZH datasets.}
% \vspace{-2mm}
% \end{figure}

\subsubsection{Cross-lingual Instruction Following or Multilingual Instruction Tuning?}
\label{sec:p_src}
As shown in Figure \ref{fig:cif_p_cross}, multilingual instruction tuning ($p_{\textit{src}}=1$) is inferior to cross-lingual instruction following ($p_{\textit{src}}<1$) for the models evaluated. 
Moreover, the result becomes suboptimal when all samples are transferred into the cross-lingual format ($p_{\textit{src}}=0$). 
We empirically set the $p_{\textit{src}}$ to 0.5 in the cross-lingual instruction following task.

\begin{table*}[thp]

\renewcommand\arraystretch{1.1}

\centering
\scriptsize

\setlength{\tabcolsep}{0.8mm}

 \begin{tabu}{l|c|c|c|c|c|c|c|c|c|c|c}
 
 \toprule[1.2pt]
  \multicolumn{1}{c}{ } & \multicolumn{5}{c}{\textbf{XNLI}} & \multicolumn{5}{c}{\textbf{XCOPA}} &\\
  \cmidrule(r){2-6}  \cmidrule(r){7-11} \noalign{\smallskip}
  \multicolumn{1}{c}{ } & \multicolumn{2}{c}{\textbf{High}} & \multicolumn{2}{c}{\textbf{Medium}} & \multicolumn{1}{c}{\textbf{Low}} & \multicolumn{2}{c}{\textbf{High}} & \multicolumn{2}{c}{\textbf{Medium}} & \multicolumn{1}{c}{\textbf{Low}} &\\
  \cmidrule(r){2-3} \cmidrule(r){4-5} \cmidrule(r){6-6} \cmidrule(r){7-8} \cmidrule(r){9-10} \cmidrule(r){11-11} \noalign{\smallskip}
\multicolumn{1}{c}{\textbf{Model}}& \textbf{EN-0/5}&\textbf{ZH-0/5}& \textbf{TH${}^{\dagger}$-0/5}&\textbf{TR${}^{\dagger}$-0/5}& \textbf{SW-0/5}&\textbf{EN-0/5}& \textbf{ZH-0/5}&\textbf{TH${}^{\dagger}$-0/5} & \textbf{TR${}^{\dagger}$-0/5}&\textbf{SW-0/5} &\multicolumn{1}{c}{\textbf{Avg}}\\

   \midrule[0.8pt]

 $\text{GPT-3}_{\text{6.7B}}$                  & $55.3/52.8$                      & $42.4/45.9$                       & $38.5/36.6$           & $40.5/38.4$                   & $34.8/33.9$                   & $73.6/74.5$                   & $55.0/57.7$   & $53.7/54.4$   & $53.4/53.0$   & $52.3/52.1$ & $49.9$\\
  \midrule[0.8pt]
 
  $\text{XGLM}_{\text{564M}}$                  & $45.5/41.2$                      & $37.6/35.6$                       & $40.8/35.0$                   & $40.2/34.9$                   & $37.5/34.7$                   & $56.4/59.6$                   & $52.8/52.2$   & $55.4/54.2$   & $52.8/51.8$   & $51.8/51.6$& $46.1$\\

 \ \ \ \ \ \ \ \ \ +MIT            & $46.8/43.4$    & $40.3/39.8$     & $41.4/39.6$ & $40.2/36.7$ & $37.6/37.9$ & $58.0/60.2$ & $55.2/55.8$   & $56.8/57.4$   & $55.4/54.6$   & $53.2/53.8$& $48.2$\\
 
 \ \ \ \ \ \ \ \ \ +AFP            & $\textbf{48.0}/\textbf{46.3}$    & $\textbf{42.8}/\textbf{42.7}$     & $\textbf{42.8}/\textbf{43.3}$ & $\textbf{40.4}/\textbf{42.9}$ & $\textbf{38.9}/\textbf{40.0}$ & $\textbf{60.6}/\textbf{61.4}$ & $\textbf{59.0}/\textbf{59.4}$   & $\textbf{59.0}/\textbf{60.0}$   & $\textbf{56.6}/\textbf{56.0}$   & $\textbf{57.6}/\textbf{55.6}$& $\textbf{50.7}$\\

 %    \cdashlinelr{1-12}

 % $\text{XGLM}_{\text{1.7B}}$                  & $47.6/42.9$                      & $39.0/39.3$                       & $40.9/37.6$           & $40.0/38.1$                   & $37.8/37.6$                   & $62.4/64.0$                   & $56.4/57.8$   & $58.4/59.4$   & $53.6/52.8$   & $54.4/52.8$& $48.6$\\
  
 % \ \ \ \ \ \ \ \ \ +AFP            & $\textbf{48.5}/\textbf{47.7}$    & $\textbf{43.6}/\textbf{44.2}$     & $\textbf{43.2}/\textbf{44.0}$ & $\textbf{42.5}/\textbf{41.8}$ & $\textbf{41.0}/\textbf{40.4}$ & $\textbf{65.4}/\textbf{66.2}$ & $\textbf{60.8}/\textbf{61.6}$   & $\textbf{62.2}/\textbf{60.4}$   & $\textbf{57.6}/\textbf{57.0}$   & $\textbf{59.2}/\textbf{59.6}$& $\textbf{52.3}$\\

    \cdashlinelr{1-12}

  $\text{XGLM}_{\text{7.5B}}$                  & $54.1/49.9$                      & $45.4/44.2$                       & $45.2/43.6$           & $44.7/39.5$                   & $44.3/39.6$                   & $69.4/74.6$                   & $62.4/63.2$   & $62.0/62.4$   & $56.6/58.4$   & $58.2/57.2$& $53.7$\\
 \ \ \ \ \ \ \ \ \ +MIT            & $54.6/51.3$    & $47.2/46.4$     & $46.5/45.7$ & $45.9/41.6$ & $45.0/41.3$ & $70.6/74.4$ & $64.0/65.2$   & $62.8/63.2$   & $58.0/59.6$   & $58.8/58.4$& $55.0$\\

 \ \ \ \ \ \ \ \ \ +AFP            & $\textbf{55.8}/\textbf{54.1}$    & $\textbf{50.6}/\textbf{48.8}$     & $\textbf{48.1}/\textbf{47.2}$ & $\textbf{46.7}/\textbf{44.1}$ & $\textbf{46.1}/\textbf{44.2}$ & $\textbf{71.4}/\textbf{75.0}$ & $\textbf{66.8}/\textbf{66.6}$   & $\textbf{63.2}/\textbf{64.4}$   & $\textbf{61.8}/\textbf{62.0}$   & $\textbf{62.2}/\textbf{62.8}$& $\textbf{57.1}$\\
 
 \specialrule{0em}{0pt}{0pt}

    % \cdashlinelr{1-9}
  \midrule[0.8pt]
  
$\text{BLOOMZ}_{\text{560M}}$                  & $43.8/44.5$                      & $41.5/40.7$                       & $37.8/39.2$                   & $35.6/35.9$                   & $35.8/35.8$                   & $54.8/57.2$                   & $52.0/52.8$   & $52.6/52.5$   & $52.6/51.8$   & $52.0/52.4$& $46.1$\\
  
$\text{BLOOM}_{\text{560M}}$                  & $44.4/40.4$                      & $41.1/40.3$                       & $33.4/35.1$                   & $34.5/34.1$                   & $35.7/34.5$                   & $53.0/57.4$                   & $49.8/54.0$   & $50.8/51.8$   & $52.8/52.6$   & $51.2/52.0$& $44.9$\\
  
 \ \ \ \ \ \ \ \ \ +AFP           & $\textbf{48.4}/\textbf{46.5}$    & $\textbf{47.4}/\textbf{44.1}$     & $\textbf{39.8}/\textbf{40.5}$ & $\textbf{39.7}/\textbf{39.4}$ & $\textbf{40.1}/\textbf{40.8}$ & $\textbf{56.0}/\textbf{58.4}$ & $\textbf{52.4}/\textbf{54.4}$   & $\textbf{53.8}/\textbf{53.4}$   & $\textbf{54.6}/\textbf{54.8}$   & $\textbf{52.2}/\textbf{53.4}$& $\textbf{48.5}$\\

    \cdashlinelr{1-12}

$\text{BLOOMZ}_{\text{1.7B}}$                  & $50.3/51.2$                      & $48.0/46.2$                       & $38.4/36.8$                   & $37.1/37.4$                  & $38.3/38.7$                   & $58.0/58.0$                   & $55.2/56.8$   & $52.4/53.8$   & $52.2/54.6$   & $50.8/50.2$& $48.2$\\

$\text{BLOOM}_{\text{1.7B}}$                  & $50.4/44.4$                      & $47.6/46.1$                       & $37.9/35.7$                   & $36.9/35.0$                   & $36.3/36.7$                   & $55.8/58.2$                   & $52.4/54.6$   & $51.2/52.0$   & $53.4/54.2$   & $52.2/53.6$& $47.2$\\

 \ \ \ \ \ \ \ \ \ +AFP           & $\textbf{52.1}/\textbf{51.3}$    & $\textbf{49.1}/\textbf{47.1}$     & $\textbf{41.2}/\textbf{41.8}$ & $\textbf{40.1}/\textbf{41.3}$ & $\textbf{41.1}/\textbf{42.5}$ & $\textbf{60.2}/\textbf{60.4}$ & $\textbf{55.4}/\textbf{58.8}$   & $\textbf{54.2}/\textbf{54.6}$   & $\textbf{55.6}/\textbf{56.0}$   & $\textbf{53.6}/\textbf{55.0}$& $\textbf{50.6}$\\
 
    \cdashlinelr{1-12}

 $\text{BLOOMZ}_{\text{7.1B}}$                  & $51.1/52.0$                      & $49.7/48.0$                       & $40.9/37.6$                   & $39.8/36.1$                  & $39.2/39.7$                   & $61.2/62.4$                   & $57.6/59.8$   & $53.2/51.6$   & $55.0/54.2$   & $53.6/52.2$& $49.7$\\

$\text{BLOOM}_{\text{7.1B}}$                  & $54.0/48.7$                      & $48.1/47.5$                       & $39.5/37.4$                   & $38.2/35.0$                   & $37.7/38.9$                   & $58.0/58.8$                   & $54.0/54.8$   & $52.6/52.8$   & $53.8/53.4$   & $53.2/54.6$& $48.6$\\

 \ \ \ \ \ \ \ \ \ +AFP           & $\textbf{55.7}/\textbf{52.5}$    & $\textbf{50.1}/\textbf{50.2}$     & $\textbf{43.7}/\textbf{43.2}$ & $\textbf{43.0}/\textbf{43.4}$ & $\textbf{42.2}/\textbf{43.1}$ & $\textbf{62.6}/\textbf{62.8}$ & $\textbf{58.2}/\textbf{60.4}$   & $\textbf{55.6}/\textbf{55.2}$   & $\textbf{56.4}/\textbf{56.6}$   & $\textbf{55.0}/\textbf{55.8}$& $\textbf{52.3}$\\
 
 \specialrule{0em}{0pt}{0pt}

\bottomrule[1.2pt]
\end{tabu}
\vspace{-2mm}

\caption{\label{tab:5lang} In-context learning performance on NLI and Reasoning datasets across 5 languages. ``\textbf{High}'', ``\textbf{Medium}'' and ``\textbf{Low}'' denotes the available amount of linguistic resources. ${}^{\dagger}$ denotes the unseen language in the pre-training corpus of BLOOM. Following \citet{lin-etal-2022-shot}, the prompt template is written in English for all languages evaluated. 
}

\vspace{-2mm}

\end{table*}
\begin{table*}[ht]

\renewcommand\arraystretch{0.9}

\centering
\footnotesize

\setlength{\tabcolsep}{2.9mm}

 \begin{tabu}{c|lccccc|c}
 
 \toprule[1.2pt]

\multicolumn{1}{c}{\ } & \multicolumn{1}{c}{\textbf{Model}}& \textbf{EN}&\textbf{ZH}& \textbf{TH}&\textbf{TR}& \multicolumn{1}{c}{\textbf{SW}} & \textbf{Avg} \\

   \midrule[0.8pt]
   
 \multirow{4}{*}{\specialcell{\textbf{Avg translate from}\\\textbf{the language}}}
&$\text{XGLM}_{\text{564M}}$          & $4.8_{\pm 0.8}$             & $2.1_{\pm 1.7}$                   & $2.2_{\pm 1.8}$           & $2.2_{\pm 1.8}$                   & $1.1_{\pm 0.9}$               & $2.5_{\pm 1.7}$ \\

& \ \ \ \ \ \ \ \ \ +AFP             & $\textbf{5.2}_{\pm 0.4}$     & $\textbf{2.7}_{\pm 1.5}$          & $\textbf{2.8}_{\pm 1.7}$  & $\textbf{2.7}_{\pm 1.6}$          & $\textbf{2.6}_{\pm 0.8}$      & $\textbf{3.2}_{\pm 1.4}$ \\
    \cdashlinelr{2-8}

& $\text{XGLM}_{\text{7.5B}}$        & $16.3_{\pm 2.2}$                      & $10.0_{\pm 6.5}$                   & $11.0_{\pm 7.0}$               & $10.0_{\pm 7.3}$                   & $14.0_{\pm 9.8}$      & $12.3_{\pm 7.5}$ \\

& \ \ \ \ \ \ \ \ \ +AFP             & $\textbf{16.9}_{\pm 1.7}$             & $\textbf{11.1}_{\pm 6.0}$          & $\textbf{11.6}_{\pm 6.8}$      & $\textbf{11.1}_{\pm 6.9}$          & $\textbf{14.7}_{\pm 9.2}$        & $\textbf{13.1}_{\pm 7.0}$ \\
 
 \specialrule{0em}{0pt}{0pt}

    % \cdashlinelr{1-9}
  \midrule[0.8pt]
\multirow{4}{*}{\specialcell{\textbf{Avg translate to}\\\textbf{the language}}}
&$\text{XGLM}_{\text{564M}}$            & $7.4_{\pm 0.7}$            & $1.3_{\pm 1.0}$              & $1.4_{\pm 1.1}$                  & $1.4_{\pm 0.9}$               & $0.9_{\pm 0.7}$                      & $2.5_{\pm 1.7}$ \\

& \ \ \ \ \ \ \ \ \ +AFP                & $\textbf{8.4}_{\pm 0.3}$   & $\textbf{1.6}_{\pm 0.8}$     & $\textbf{1.8}_{\pm 1.1}$         & $\textbf{2.3}_{\pm 0.8}$      & $\textbf{1.9}_{\pm 0.6}$             & $\textbf{3.2}_{\pm 1.4}$ \\
    \cdashlinelr{2-8}

& $\text{XGLM}_{\text{7.5B}}$           & $24.3_{\pm 3.3}$           & $8.0_{\pm 2.9}$              & $11.0_{\pm 4.3}$                 & $9.6_{\pm 3.7}$               & $8.5_{\pm 5.7}$                      & $12.3_{\pm 7.5}$ \\

& \ \ \ \ \ \ \ \ \ +AFP                & $\textbf{24.4}_{\pm 3.0}$  & $\textbf{10.0}_{\pm 2.2}$    & $\textbf{11.8}_{\pm 3.9}$        & $\textbf{10.3}_{\pm 3.2}$     & $\textbf{9.0}_{\pm 5.4}$             & $\textbf{13.1}_{\pm 7.0}$ \\

 \specialrule{0em}{0pt}{0pt}

\bottomrule[1.2pt]
\end{tabu}
\vspace{-2mm}

\caption{\label{tab:5langTrans} Few-shot multilingual machine translation results of spBLEU on FLORES-101 devtest set. The variance of performance across the input or output languages is marked in the subscript.}

\vspace{-4mm}

\end{table*}

\subsection{Multilingual Results and Analyses}
\label{sec:mu_res}
% en-zh-ar-ur of xglm/bloom on xnli
% en-es-ru-zh-ar-hi-sw of xglm/bloom on XNLI and XCOPA
% en-zh-th-tr-sw of xglm/bloom on XNLI and XCOPA % en & zh: high, th & tr: medium, sw: low
In addition to the bilingual alignment, AFP can be applied to align the models in multilingual conditions. 
English is first chosen as the pivot language of alignment for the dominance performance in multilingual generative models. 
That is, the input parallel samples of AFP are selected from the EN-XX corpus, e.g., EN-ZH and EN-TH, to pull the representations and outputs of models in other languages closer to the ones in English. 
We also investigate the other alignment methods like pair-wise alignment in Section \ref{Sec:eng_bridge_and_pairwise}, which shows an inferior performance. 

\begin{table}[th]
\renewcommand\arraystretch{0.8}

\centering
\footnotesize

\setlength{\tabcolsep}{1.2mm}
\begin{tabular}{lcccccc}
    \toprule[1.2pt]  
    \multicolumn{1}{c}{\textbf{Model}}                & \textbf{EN}          &  \textbf{ZH}          &  \textbf{TH}          &  \textbf{TR}          &  \textbf{SW}          &  \textbf{Avg}         \\
    \midrule[0.8pt]
    \multicolumn{1}{l}{$\text{XGLM}_{\text{564M}}$}   & $50.7$              & $44.6$               & $46.4$               & $44.9$               & $43.9$               & $46.1$               \\
        \cdashlinelr{1-7}
        {\scriptsize w/ EN as pivot language}               & $\textbf{54.1}$     & $\textbf{51.0}$      & $\textbf{51.3}$     & $49.2$                & $48.0$               & $\textbf{50.7}$      \\
        {\scriptsize w/ Pairwise alignment}           & $52.7$              & $50.5$               & $50.4$              & $\textbf{49.5}$       & $\textbf{48.4}$      & $50.3$               \\
    \midrule[0.8pt]
    \multicolumn{1}{l}{$\text{BLOOM}_{\text{560M}}$}  & $48.8$              & $46.3$               & $42.8$               & $43.5$               & $43.4$               & $45.0$               \\
        \cdashlinelr{1-7}
        {\scriptsize w/ EN as pivot language}               & $\textbf{52.3}$    & $\textbf{49.6}$       & $\textbf{46.9}$     & $\textbf{47.1}$      & $46.6$               & $\textbf{48.5}$      \\
        {\scriptsize w/ Pairwise alignment}           & $51.6$             & $48.9$                & $46.5$              & $46.3$               & $\textbf{46.8}$      & $48.0$               \\
        \bottomrule[1.2pt]
\end{tabular}
\caption{\label{tab:ablation_study_multilingual}Results of different alignment policies. The policy adopting English as pivot language achieves higher improvement on average and is adopted as default.}
\vspace{-0.2cm}
	
\end{table}
\begin{table}[th]
\renewcommand\arraystretch{1.2}

\centering
\footnotesize

\setlength{\tabcolsep}{1mm}
\begin{tabular}{lcccccc}
    \toprule[1.2pt]  
    \multicolumn{1}{c}{\textbf{Model}}                & \textbf{EN}          &  \textbf{ZH}          &  \textbf{TH}          &  \textbf{TR}          &  \textbf{SW}          &  \textbf{Avg}         \\
    \midrule[0.8pt]
    \multicolumn{1}{l}{$\text{XGLM}_{\text{7.5B}}$ + AFP}     & $64.1$              & $58.2$               & $55.7$               & $53.7$               & $53.8$               & $57.1$               \\
        \cdashlinelr{1-7}
        {\scriptsize w/ Semantic aligned demos}               & $\textbf{64.4}$     & $\textbf{58.7}$      & $\textbf{55.8}$      & $\textbf{55.8}$      & $\textbf{54.0}$      & $\textbf{57.7}$      \\
    \midrule[0.8pt]
    \multicolumn{1}{l}{$\text{BLOOM}_{\text{7.1B}}$ + AFP}    & $58.4$              & $54.7$               & $49.4$               & $49.9$               & $49.0$               & $52.3$               \\
        \cdashlinelr{1-7}
        {\scriptsize w/ Semantic aligned demos}               & $\textbf{58.9}$     & $\textbf{54.8}$      & $\textbf{49.5}$      & $\textbf{50.2}$      & $\textbf{49.2}$      & $\textbf{52.5}$      \\
        \bottomrule[1.2pt]
\end{tabular}
\caption{\label{tab:combination_semantic} The average performance on XNLI and XCOPA when prompt with 5 semantic aligned demos.}
\vspace{-0.4cm}
	
\end{table}
\begin{table*}[htp]

\renewcommand\arraystretch{1.1}

\centering
\scriptsize

\setlength{\tabcolsep}{1.2mm}

 \begin{tabu}{l|cc|cc|cc|cc|cc|c}
 
 \toprule[1.2pt]
  \multicolumn{1}{c}{\textbf{ }} & \multicolumn{2}{c}{\textbf{XNLI}} & \multicolumn{2}{c}{\textbf{PAWS-X}} & \multicolumn{2}{c}{\textbf{XCOPA}} & \multicolumn{2}{c}{\textbf{XStoryCloze}} & \multicolumn{2}{c}{\textbf{XWinograd}} & \\
  \cmidrule(r){2-3}  \cmidrule(r){4-5} \cmidrule(r){6-7} \cmidrule(r){8-9} \cmidrule(r){10-11} \noalign{\smallskip}
\multicolumn{1}{c}{\textbf{Model}}& \multicolumn{1}{c}{\textbf{0-shot}}&\multicolumn{1}{c}{\textbf{5-shot}}& \multicolumn{1}{c}{\textbf{0-shot}}&\multicolumn{1}{c}{\textbf{5-shot}}& \multicolumn{1}{c}{\textbf{0-shot}}&\multicolumn{1}{c}{\textbf{5-shot}}& \multicolumn{1}{c}{\textbf{0-shot}}&\multicolumn{1}{c}{\textbf{5-shot}} & \multicolumn{1}{c}{\textbf{0-shot}}&\multicolumn{1}{c}{\textbf{5-shot}}& \multicolumn{1}{c}{\textbf{Avg}} \\

   \midrule[0.8pt]

  $\text{XGLM}_{\text{7.5B}}$      & $45.6_{\pm 3.4}$             & $43.6_{\pm 3.1}$                & $54.7_{\pm 3.1}$                   & $55.1_{\pm 1.6}$                   & $58.9_{\pm 5.0}$                   & $60.4_{\pm 5.7}$                   & $60.6_{\pm 3.9}$   & $60.5_{\pm 5.0}$   & $63.9_{\pm 5.1}$   & $64.7_{\pm 4.2}$ & $55.3_{\pm 8.5}$ \\
  
 \ \ \ \ \ \ \ \ \ +AFP            & $\textbf{47.5}_{\pm 3.3}$    & $\textbf{47.7}_{\pm 3.0}$       & $\textbf{57.7}_{\pm 2.3}$          & $\textbf{57.5}_{\pm 1.4}$          & $\textbf{61.3}_{\pm 4.5}$          & $\textbf{62.4}_{\pm 5.7}$  & $\textbf{62.4}_{\pm 3.7}$   & $\textbf{63.5}_{\pm 4.9}$   & $\textbf{65.5}_{\pm 4.8}$   & $\textbf{66.7}_{\pm 4.1}$ & $\textbf{57.8}_{\pm 8.0}$\\
 
 \specialrule{0em}{0pt}{0pt}

  \midrule[0.8pt]
 
 $\text{BLOOMZ}_{\text{7.1B}}$     & $44.1_{\pm 4.0}$                      & $43.5_{\pm 4.6}$                       & $57.8_{\pm 2.6}$                   & $\textbf{56.6}_{\pm 2.9}$                   & $53.1_{\pm 5.3}$                   & $54.6_{\pm 5.5}$                   & $58.9_{\pm 6.7}$   & $61.0_{\pm 7.4}$   & $60.0_{\pm 4.9}$   & $60.4_{\pm 5.9}$ & $54.2_{\pm 8.3}$ \\
 $\text{BLOOM}_{\text{7.1B}}$      & $43.3_{\pm 5.5}$                      & $42.5_{\pm 4.7}$                       & $54.5_{\pm 3.1}$                   & $53.5_{\pm 3.6}$                   & $52.3_{\pm 4.7}$                   & $53.3_{\pm 4.0}$                   & $57.3_{\pm 6.2}$   & $59.2_{\pm 7.2}$   & $59.0_{\pm 6.2}$   & $59.2_{\pm 5.2}$ & $52.0_{\pm 8.2}$ \\

 \ \ \ \ \ \ \ \ \ +AFP            & $\textbf{45.4}_{\pm 4.5}$              & $\textbf{45.9}_{\pm 3.9}$             & $\textbf{58.1}_{\pm 2.6}$          & $56.1_{\pm 3.1}$          & $\textbf{55.0}_{\pm 3.7}$          & $\textbf{55.1}_{\pm 3.9}$  & $\textbf{61.3}_{\pm 6.0}$   & $\textbf{62.5}_{\pm 7.2}$   & $\textbf{61.1}_{\pm 5.8}$   & $\textbf{60.5}_{\pm 5.2}$ & $\textbf{54.7}_{\pm 8.0}$\\
 
 % \specialrule{0em}{0pt}{0pt}

 %  \midrule[0.8pt]

 % $\text{Llama}_{\text{7B}}$                  & $46.2_{\pm 3.1}$                      & $46.2_{\pm 3.1}$                       & $46.2_{\pm 3.1}$                   & $46.2_{\pm 3.1}$                   & $46.2_{\pm 3.1}$                   & $46.2_{\pm 3.1}$                   & $46.2_{\pm 3.1}$   & $46.2_{\pm 3.1}$   & $46.2_{\pm 3.1}$   & $46.2_{\pm 3.1}$ & $46.2_{\pm 3.1}$ \\

 % \ \ \ \ \ \ \ \ \ +AFP           & $\textbf{55.0}_{\pm 3.1}$    & $\textbf{55.0}_{\pm 3.1}$       & $\textbf{55.0}_{\pm 3.1}$ & $\textbf{55.0}_{\pm 3.1}$  & $\textbf{55.0}_{\pm 3.1}$ & $\textbf{55.0}_{\pm 3.1}$  & $\textbf{55.0}_{\pm 3.1}$   & $\textbf{55.0}_{\pm 3.1}$   & $\textbf{55.0}_{\pm 3.1}$   & $\textbf{55.0}_{\pm 3.1}$ & $\textbf{55.0}_{\pm 3.1}$\\

\bottomrule[1.2pt]
\end{tabu}

\vspace{-2mm}

\caption{\label{tab:5datasets_all} In-context learning results of models on 5 datasets across all languages. The variance of performance across languages is marked in the subscript. All results are reported in Appendix \ref{appendix:all_langs}. 
}

\vspace{-4mm}

\end{table*}

Table \ref{tab:5lang} reports the results of alignment between 5 languages from different language families, where the performance of models on the NLI and reasoning tasks is improved by 3.72\% from high-resource languages to the less-represented language Swahili. 

Moreover, models with AFP obtain a more balanced performance distribution. 
Taking XGLM models as an example, the variance of performance across 5 languages decreases from 3.44\% to 2.96\% on average. 
It is noted that AFP advances the performance of BLOOM in the two unseen languages, Thai (TH, +3.9\%) and Turkish (TR, +3.92\%).

Multilingual generative models also obtain a performance gain (+0.75 BLEU) in the multilingual machine translation task after alignment (Table \ref{tab:5langTrans}). 
It can also find a more balanced performance distribution across languages, where the average variance reduction is 0.4\% for the models evaluated. 

% en-zh-ar-hi-sw (eng-zho\_simple-ara-hin-swh) FLORES-101 translation results of xglm-7b and bloom-7b

\subsubsection{English as a pivot language or Pairwise Alignment?}
\label{Sec:eng_bridge_and_pairwise}
Besides adopting English as a pivot language to align multilingual representations, we also investigate the pairwise alignment policy, which is aligned by languages in pairs. 
For example, assuming to align the representations of English (EN), Chinese (ZH), and Thai (TH), the former policy comes to two parallel samples for input, which are EN-ZH and EN-TH, while the latter contains three parallel samples: EN-ZH, EN-TH, and ZH-TH. 

The results of five languages alignment experiments on XNLI and XCOPA are reported in Table \ref{tab:ablation_study_multilingual}. 
The pairwise alignment policy performs consistently better in the low-resource language Swahili, although its average improvement is inferior to that when adopting English as a pivot language. 

\subsubsection{Combination with Other Cross-lingual Methods}
After alignment, multilingual generative models can use other cross-lingual methods for further improvement. 
We take a method named semantic alignment for an example, which is able to promote the cross-lingual ability using semantic aligned demos in prompt \citep{tanwar-etal-2023-multilingual}. 
As shown in Table \ref{tab:combination_semantic}, models obtain a further 0.4\% improvement in the multilingual NLI and reasoning tasks on average.

\subsection{Extended to Alignment in 52 Languages}
Based on the above analyses, we extend the alignment to all 52 languages in the Bactrian-X dataset by adopting English as a pivot language (information about all languages involved is reported in Appendix \ref{appendix:lang_code}). 
As shown in Table \ref{tab:5datasets_all}, models obtain a 2.6\% improvement in 5 multilingual tasks on average, and mitigate the variance across languages. 
It is also noted that the performance of $\text{BLOOM}_{\text{7.1B}}$ on unseen languages among 5 datasets is improved by 2.8\% using only parallel samples via our alignment framework, which may arise from the knowledge transferred from other languages after alignment. 
% Report the average performance on XNLI, PAWS-X, XStoryCloze, XCOPA and XWinograd performance on all languages.

\subsection{Ablation Study}
\label{sec:ablation}
To take a deep look into the improvements contributed by AFP, we conduct an ablation study on the 5 datasets of bilingual tasks using $\text{XGLM}_{\text{564M}}$ (Table \ref{tab:ablation_study_xglm_564m}). 

The in-context learning abilities of the models decrease when only multilingual contrastive learning (MCL) is used. 
It may arise from the next word prediction ability of the model is affected by the MCL. 
Using the same data, both multilingual instruction tuning (MIT, +1.3\%) and cross-lingual instruction following (CIF, +2.1\%) can improve multilingual generative models, while the latter can promote it more. 
In addition, the performance of the models can be further improved after combining MCL and CIF, which is the proposed alignment framework AFP. 

\begin{table}[t]
\vspace{1mm}
        \setlength{\tabcolsep}{3mm}
	\centering
	\scriptsize
	%\vspace{-0.4cm}
	\renewcommand\arraystretch{1.2}
	\begin{center}
		% \caption{Ablation study of different training methods on 5 datasets for $\text{XGLM}_{\text{564M}}$.}
		\begin{tabular}{lccc}
			\toprule[1.2pt]  
			\multicolumn{1}{c}{\textbf{Model}}                & \textbf{0-shot}              &  \textbf{3-shot}              &  \textbf{5-shot} \\
			\midrule[0.8pt]
			\multicolumn{1}{l}{$\text{XGLM}_{\text{564M}}$}   & $52.94_{\pm 0.54}$           & $51.71_{\pm 0.90}$            & $52.03_{\pm 0.89}$\\
                \midrule[0.8pt]
                {\scriptsize w/ MCL}                              & $50.23_{\pm 0.43}$           & $48.66_{\pm 0.51}$            & $48.60_{\pm 0.49}$\\
                % {\scriptsize w/ CLM}                              & $54.51_{\pm 0.57}$           & $53.27_{\pm 0.77}$            & $52.60_{\pm 0.52}$\\
                {\scriptsize w/ MIT}                              & $54.25_{\pm 0.49}$           & $53.54_{\pm 0.75}$            & $52.93_{\pm 0.68}$\\
                {\scriptsize w/ CIF}                              & $55.31_{\pm 0.55}$           & $54.02_{\pm 0.63}$            & $53.58_{\pm 0.64}$\\
			{\scriptsize w/ AFP}                              & $\textbf{55.97}_{\pm 0.48}$  & $\textbf{55.50}_{\pm 0.55}$   & $\textbf{56.15}_{\pm 0.43}$\\
			\bottomrule[1.2pt]
		\end{tabular}
	\end{center}
	\vspace{-2mm}
        \caption{\label{tab:ablation_study_xglm_564m}Ablation study of different training methods on 5 datasets for $\text{XGLM}_{\text{564M}}$.}
	\vspace{-4mm}
\end{table}

\section{Conclusion and Future Work}
In this paper, we proposed a simple yet effective multilingual alignment framework, including internal multilingual representations alignment and cross-lingual outputs alignment methods. 
Experimental results show that this framework improves both the internal representations and cross-lingual capabilities of generative models across various scales. 

Beyond aligning different languages, our framework can be extended to align the internal representations and outputs across different modalities in the multi-modal generative models by replacing parallel samples. 
However, it is noted that the current framework relies on labeled training data for alignment.
Future works can focus on the unsupervised multilingual alignment method for language models. 

\section*{Limitations}
% These are limitations of our work. 
Firstly, although our cross-lingual framework boosts the cross-lingual ability of multilingual generative language models using only a small amount of parallel samples, it is noted that the proposed framework relies on labeled training data for alignment, which is unavailable for languages without parallel samples. 

In addition, due to limited computation resources, our framework is constrained to multilingual generative language models with less than or equal to 7.5B parameters. 

Lastly, there is an error propagation problem from the involved machine translation system, which may result in inferior performance. 

\section*{Ethical Considerations}
Since our alignment framework is applied to the pre-trained multilingual generative language models, the model aligned may inherit the potential risk and bias in the vanilla language model \citep{tamkin2021understanding}. 
The cultural bias and offensive response in English may be incorporated into other languages due to the alignment policy used, which adopts English as a pivot language. 
Future explorations are needed to mitigate the risk and cultural bias in multilingual generative language models. 

\section*{Acknowledgements}
We thank anonymous reviewers for their insightful comments and suggestions. 
The research work was supported by the National Science Foundation of China (No. 62036001 and No. 62122088). 

\bibliography{anthology,custom}

\begin{figure*}[ht]
% \flushleft
\centering
\subfigure[]{\includegraphics [scale=0.5]{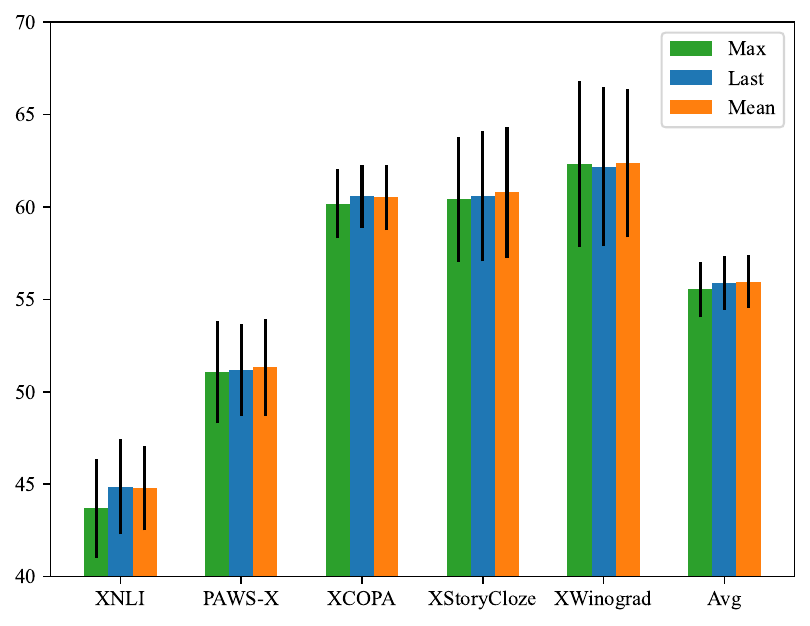}\label{fig:pooling}}
\subfigure[]{\includegraphics [scale=0.238]{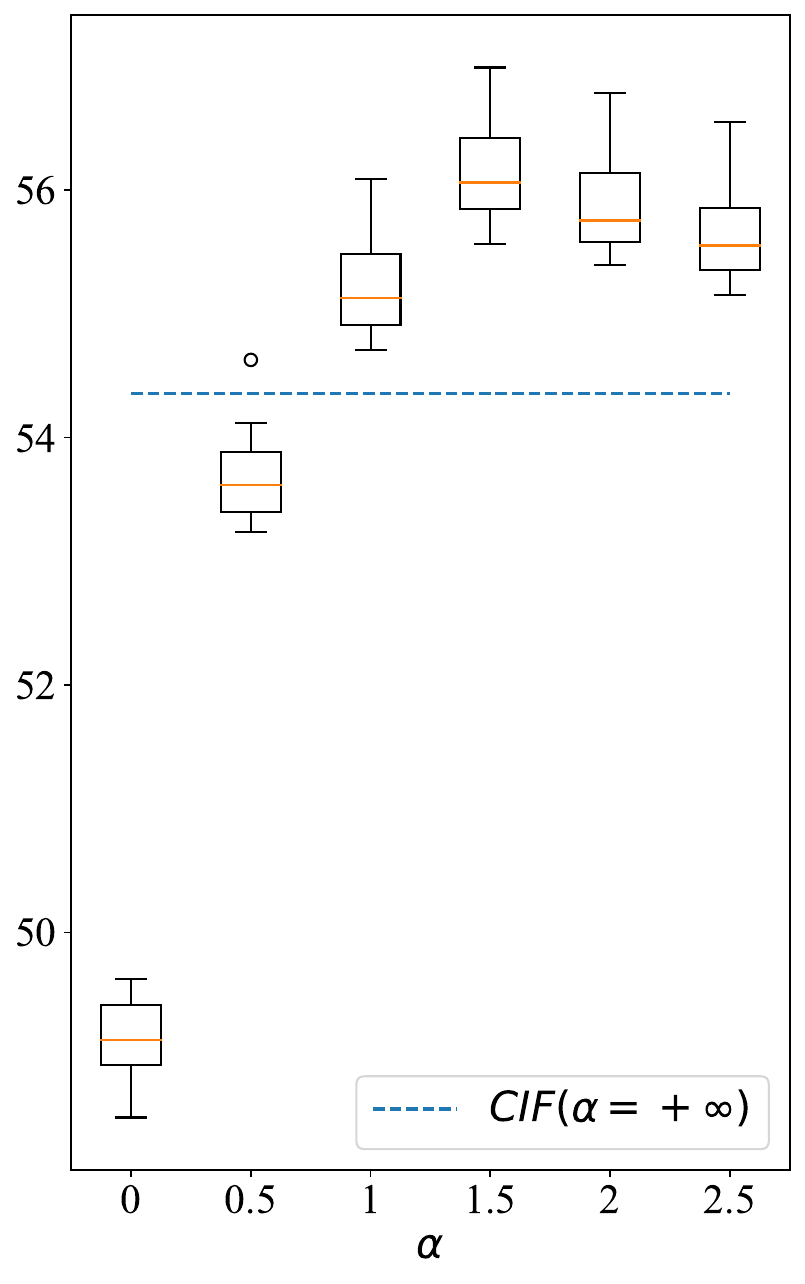}\label{fig:cif_weight}}
\caption{Results of different pooling methods (a) and weights of CIF (b) on 5 EN-ZH datasets using $\text{XGLM}_{\text{564M}}$. }
\vspace{-2mm}
\end{figure*}

\appendix
\section{Hyperparameters}
\label{appendix:hyper}
To align the representations and outputs of multilingual generative models, we adopt AdamW \citep{loshchilov2019adamw} optimizer, where $\beta_1 = 0.9$ and $\beta_2 = 0.999$, and a learning rate of 1e-5. 
The temperature $\tau$ is set to 0.05 in the multilingual contrastive learning task. 
Mixed precision training and ZeRO are applied to speed up the training process and save memory used \citep{micikevicius2018mixed, rasley2020deepspeed}. 
The number of training steps is empirically set to 10k with a batch size of 128. 
All experiments are conducted on a GPU server with 8*A100 80GB RAM.

\section{Additional Results}

\subsection{Pooling Methods}
Given representations for each token in the sentence, there are three general methods, the last token representation, max pooling and mean pooling, to obtain the representation of this sentence. 
Figure \ref{fig:pooling} illustrates the results of $\text{XGLM}_{\text{564M}}$ under different pooling methods using AFP. 
It can be found that the last token and mean pooling perform better, and our method is less sensitive to the pooling method chosen. 
Thus, these two methods are used in AFP and are selected according to the performance of the development set.

\subsection{Weight of Cross-lingual Instruction Following}
We find that the weight $\alpha$ of cross-lingual instruction following in Eq. (\ref{eq:afp_loss}) affects the multilingual performance of models. 
The average performance of $\text{XGLM}_{\text{564M}}$ on 5 datasets with different $\alpha$ is presented in Figure \ref{fig:cif_weight}, where models perform better than the other values evaluated when $\alpha$ is set to 1.5. 
Therefore, we only consider a limited hyperparameter sweep for each multilingual generative model with 
$\alpha \in \{1, 1.5, 2\}$.

\subsection{Distribution of multilingual representations}
\label{appendix:multilingual_dist}
\begin{figure*}[ht]
% \flushleft
\centering
\subfigure[]{\includegraphics [scale=0.33]{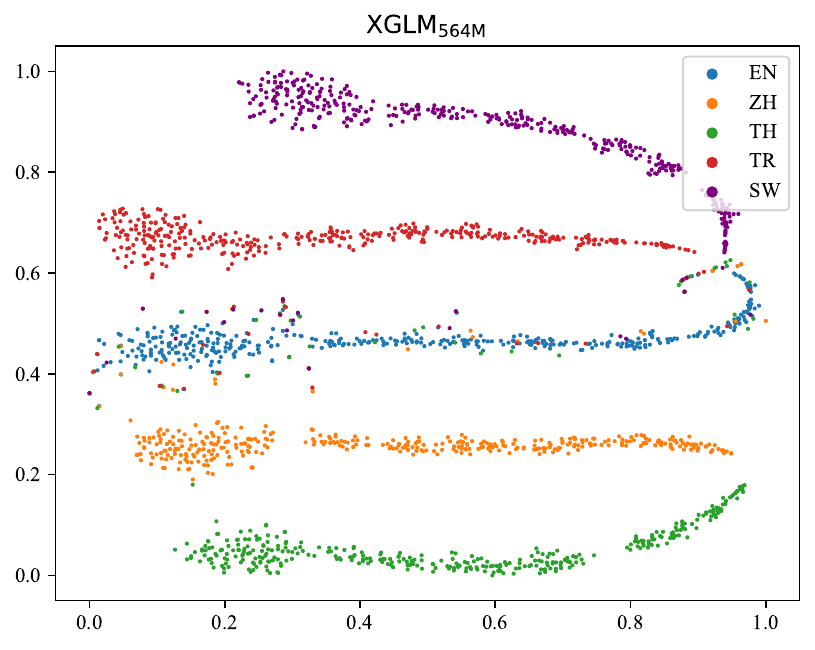}\label{fig:5langs_xglm_564m}}
\subfigure[]{\includegraphics [scale=0.33]{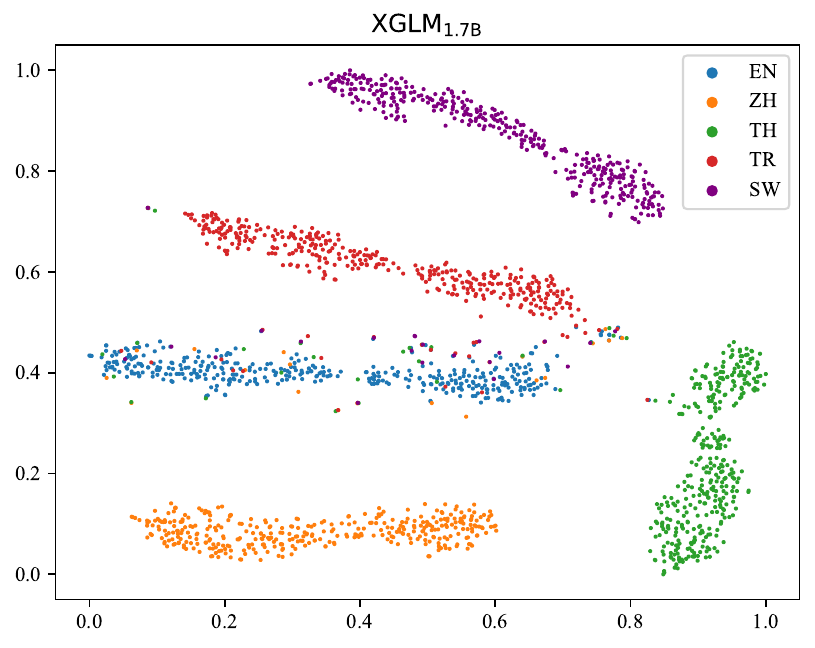}\label{fig:5langs_xglm_1.7b}}
\subfigure[]{\includegraphics [scale=0.33]{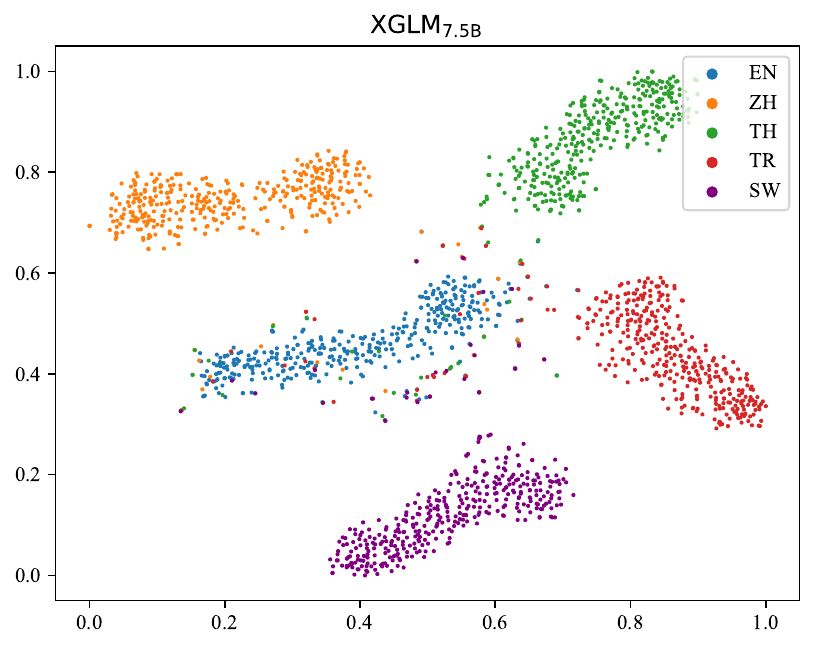}\label{fig:5langs_xglm_7.5b}}
\subfigure[]{\includegraphics [scale=0.33]{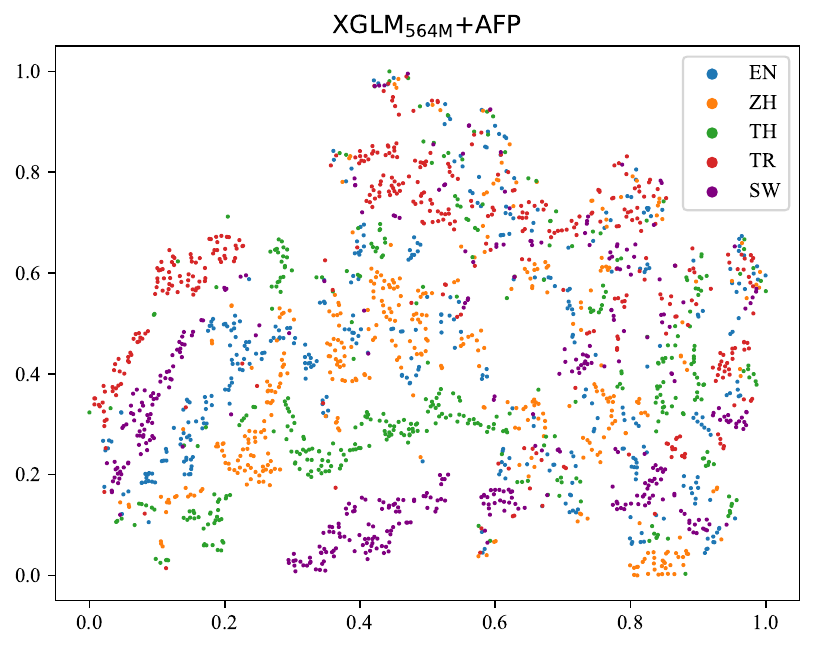}\label{fig:5langs_xglm-afp_564m}}
\subfigure[]{\includegraphics [scale=0.33]{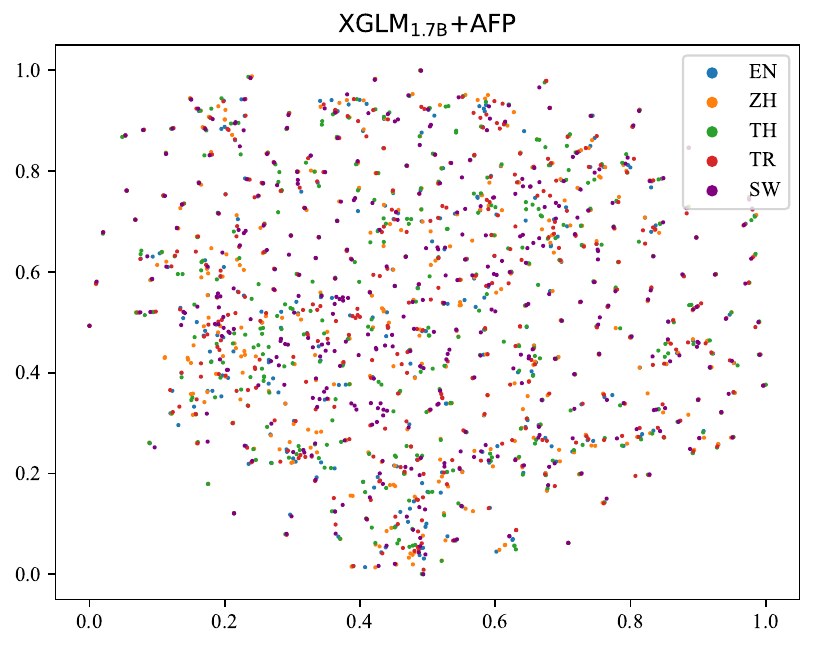}\label{fig:5langs_xglm-afp_1.7b}}
\subfigure[]{\includegraphics [scale=0.33]{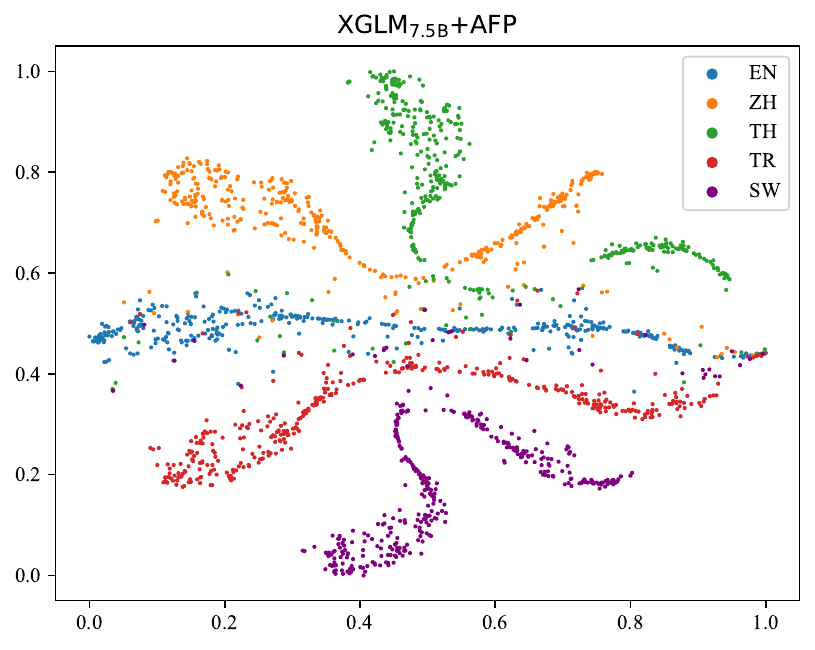}\label{fig:5langs_xglm-afp_7.5b}}
\caption{\label{fig:5langs_xglm}Distribution of multilingual sentence representations in XGLM.(Vanilla:(a)-(c), Aligned:(d)-(f), shown in t-SNE)}
\vspace{-2mm}
\end{figure*}

Figure \ref{fig:5langs_xglm} illustrates the distributions of 5 languages sentence representations from the vanilla XGLM models and the aligned ones via t-SNE. 
Similar to the bilingual distribution, we can find that there are distinct gaps between the sentence representations from different languages in the vanilla models (Figure \ref{fig:5langs_xglm_564m}-\ref{fig:5langs_xglm_7.5b}). 
After training with AFP, the multilingual sentence distributions of models are better aligned between languages across different scales (Figure \ref{fig:5langs_xglm-afp_564m}-\ref{fig:5langs_xglm-afp_7.5b}). 
The alignment of multilingual sentence representations in $\text{XGLM}_{\text{7.5B}}$ is not as good as the two smaller models, which may arise from the limited parallel samples used. 

\subsection{Performance on multilingual datasets}
\label{appendix:all_langs}
All results of $\text{XGLM}_{\text{7.5B}}$, $\text{BLOOMZ}_{\text{7.1B}}$, and $\text{BLOOM}_{\text{7.1B}}$ on the 5 multilingual datasets are reported in Tabel \ref{tab:all_xnli}-\ref{tab:all_xwinograd}. 

\begin{table*}[thp]

\renewcommand\arraystretch{1.3}

\centering
\scriptsize

\setlength{\tabcolsep}{0.85mm}

 \begin{tabu}{l|c|cccccc|cccccc|ccc|c}
 
 \toprule[1.2pt]
  \multicolumn{2}{c}{ } & \multicolumn{6}{c}{\textbf{High}} & \multicolumn{6}{c}{\textbf{Medium}} & \multicolumn{3}{c}{\textbf{Low}} &\\
  \cmidrule(r){3-8} \cmidrule(r){9-14} \cmidrule(r){15-17} \noalign{\smallskip}
\multicolumn{1}{c}{\textbf{Model}}& \multicolumn{1}{c}{\textbf{\#shot}} & EN & DE${}^{\dagger}$ & ES & FR & RU${}^{\dagger}$  & ZH & AR & BG${}^{\dagger}$  & EL${}^{\dagger}$  & TH${}^{\dagger}$  & TR${}^{\dagger}$  & VI & HI & SW & UR &\multicolumn{1}{c}{\textbf{Avg}}\\

   \midrule[0.8pt]
\multirow{2}{*}{$\text{XGLM}_{\text{7.5B}}$}
            & 0 & $54.1$  &  $42.5$  &  $39.9$  &  $49.9$  &  $45.0$  &  $45.4$  &  $46.4$  &  $48.9$  &  $45.4$  &  $45.2$  &  $44.7$  &  $47.2$  &  $43.2$  &  $44.3$  &  $42.1$  &  $45.6$\\
            & 5 & $49.9$  &  $43.1$  &  $48.5$  &  $45.8$  &  $42.5$  &  $44.2$  &  $41.9$  &  $43.8$  &  $45.6$  &  $43.6$  &  $39.5$  &  $46.1$  &  $42.2$  &  $39.6$  &  $38.2$  &  $43.6$\\
    \cdashlinelr{1-18}
\multirow{2}{*}{$\text{XGLM}_{\text{7.5B}}$ +AFP}
            & 0 & $54.8$  &  $44.6$  &  $41.9$  &  $51.3$  &  $48.4$  &  $50.6$  &  $48.8$  &  $48.8$  &  $47.2$  &  $48.6$  &  $47.4$  &  $47.7$  &  $44.4$  &  $45.3$  &  $42.6$  &  $\textbf{47.5}$\\
            & 5 & $51.7$  &  $48.2$  &  $50.7$  &  $51.0$  &  $47.1$  &  $48.9$  &  $47.5$  &  $47.3$  &  $48.8$  &  $49.3$  &  $45.5$  &  $50.2$  &  $44.9$  &  $43.7$  &  $40.2$  &  $\textbf{47.7}$\\

  \midrule[0.8pt]
\multirow{2}{*}{$\text{BLOOMZ}_{\text{7.1B}}$}
            & 0 & $51.1$  &  $43.7$  &  $41.4$  &  $48.0$  &  $42.3$  &  $49.7$  &  $48.2$  &  $40.3$  &  $39.7$  &  $40.9$  &  $39.8$  &  $48.7$  &  $45.6$  &  $39.2$  &  $42.9$  &  $44.1$\\
            & 5 & $52.0$  &  $45.3$  &  $43.2$  &  $50.2$  &  $41.6$  &  $48.0$  &  $45.9$  &  $41.3$  &  $38.0$  &  $37.6$  &  $36.1$  &  $47.7$  &  $45.4$  &  $39.7$  &  $40.1$  &  $43.5$\\
    \cdashlinelr{1-18}

\multirow{2}{*}{$\text{BLOOM}_{\text{7.1B}}$}
            & 0 & $54.0$  &  $39.2$  &  $41.5$  &  $51.7$  &  $41.3$  &  $48.1$  &  $47.4$  &  $37.8$  &  $36.3$  &  $39.3$  &  $38.9$  &  $48.9$  &  $47.4$  &  $37.7$  &  $39.9$  &  $43.3$\\
            & 5 & $48.7$  &  $43.5$  &  $42.8$  &  $50.3$  &  $39.1$  &  $47.5$  &  $45.7$  &  $40.7$  &  $35.1$  &  $37.4$  &  $35.0$  &  $48.0$  &  $44.1$  &  $38.9$  &  $40.4$  &  $42.5$\\
    \cdashlinelr{1-18}
\multirow{2}{*}{$\text{BLOOM}_{\text{7.1B}}$ +AFP}
            & 0 & $55.0$  &  $42.2$  &  $43.8$  &  $52.7$  &  $42.4$  &  $48.1$  &  $50.0$  &  $40.3$  &  $40.4$  &  $42.3$  &  $41.6$  &  $50.0$  &  $45.3$  &  $42.0$  &  $45.1$  &  $\textbf{45.4}$\\
            & 5 & $53.3$  &  $44.5$  &  $44.1$  &  $51.9$  &  $44.1$  &  $49.8$  &  $49.2$  &  $42.4$  &  $42.3$  &  $41.9$  &  $41.3$  &  $50.7$  &  $47.0$  &  $42.4$  &  $44.1$  &  $\textbf{45.9}$\\

%    \midrule[0.8pt]
% \multirow{2}{*}{$\text{Llama}_{\text{7B}}$}
%             & 0 & $55.3$     & $55.3$     & $55.3$     & $55.3$     & $55.3$     & $55.3$     & $55.3$     & $55.3$     & $55.3$     & $55.3$     & $55.3$     & $55.3$     & $55.3$     & $55.3$     & $55.3$     & $55.3$ \\
%             & 5 & $55.3$     & $55.3$     & $55.3$     & $55.3$     & $55.3$     & $55.3$     & $55.3$     & $55.3$     & $55.3$     & $55.3$     & $55.3$     & $55.3$     & $55.3$     & $55.3$     & $55.3$     & $55.3$ \\
%     \cdashlinelr{1-18}
% \multirow{2}{*}{$\text{Llama}_{\text{7B}}$ +AFP}
%             & 0 & $55.3$     & $55.3$     & $55.3$     & $55.3$     & $55.3$     & $55.3$     & $55.3$     & $55.3$     & $55.3$     & $55.3$     & $55.3$     & $55.3$     & $55.3$     & $55.3$     & $55.3$     & $55.3$ \\
%             & 5 & $55.3$     & $55.3$     & $55.3$     & $55.3$     & $55.3$     & $55.3$     & $55.3$     & $55.3$     & $55.3$     & $55.3$     & $55.3$     & $55.3$     & $55.3$     & $55.3$     & $55.3$     & $55.3$ \\
    
\bottomrule[1.2pt]
\end{tabu}

\caption{\label{tab:all_xnli} In-context learning results on XNLI across all languages. ``\textbf{High}'', ``\textbf{Medium}'' and ``\textbf{Low}'' denotes the available amount of linguistic resources. ${}^{\dagger}$ denotes the unseen language in the pre-training corpus of BLOOM.
}

\vspace{-2mm}

\end{table*}
\begin{table*}[thp]

\renewcommand\arraystretch{1.3}

\centering
\scriptsize

\setlength{\tabcolsep}{0.8mm}

 \begin{tabu}{l|c|cccccc|c|c}
 
 \toprule[1.2pt]
  \multicolumn{2}{c}{ } & \multicolumn{6}{c}{\textbf{High}} & \multicolumn{1}{c}{\textbf{Medium}} &\\
  \cmidrule(r){3-8} \cmidrule(r){9-9} \noalign{\smallskip}
\multicolumn{1}{c}{\textbf{Model}}& \textbf{\#shot} & EN & DE${}^{\dagger}$  & ES & FR & ZH & JA${}^{\dagger}$  & KO${}^{\dagger}$  &\multicolumn{1}{c}{\textbf{Avg}}\\

   \midrule[0.8pt]
\multirow{2}{*}{$\text{XGLM}_{\text{7.5B}}$}
            & 0 & $58.9$  &  $58.0$  &  $57.3$  &  $54.0$  &  $52.9$  &  $50.3$  &  $51.6$  &  $54.7$\\
            & 5 & $56.3$  &  $56.1$  &  $56.6$  &  $55.8$  &  $55.8$  &  $53.3$  &  $52.2$  &  $55.1$\\
    \cdashlinelr{1-10}
\multirow{2}{*}{$\text{XGLM}_{\text{7.5B}}$ +AFP}
            & 0 & $61.4$  &  $58.2$  &  $58.2$  &  $59.4$  &  $57.5$  &  $56.2$  &  $53.4$  &  $\textbf{57.7}$\\
            & 5 & $59.4$  &  $59.0$  &  $57.4$  &  $58.1$  &  $57.6$  &  $56.3$  &  $54.9$  &  $\textbf{57.5}$\\
    
  \midrule[0.8pt]
\multirow{2}{*}{$\text{BLOOMZ}_{\text{7.1B}}$}
            & 0 & $63.6$  &  $57.9$  &  $58.5$  &  $57.4$  &  $56.9$  &  $55.7$  &  $54.8$  &  $57.8$\\
            & 5 & $62.2$  &  $56.9$  &  $57.3$  &  $57.1$  &  $56.1$  &  $54.7$  &  $51.8$  &  $\textbf{56.6}$\\
    \cdashlinelr{1-10}

\multirow{2}{*}{$\text{BLOOM}_{\text{7.1B}}$}
            & 0 & $59.9$  &  $54.7$  &  $57.7$  &  $54.0$  &  $53.2$  &  $52.0$  &  $50.3$  &  $54.5$\\
            & 5 & $60.4$  &  $54.4$  &  $54.9$  &  $54.9$  &  $51.4$  &  $50.4$  &  $48.5$  &  $53.5$\\
    \cdashlinelr{1-10}
\multirow{2}{*}{$\text{BLOOM}_{\text{7.1B}}$ +AFP}
            & 0 & $62.4$  &  $59.3$  &  $59.6$  &  $59.2$  &  $56.2$  &  $56.1$  &  $54.2$  &  $\textbf{58.1}$\\
            & 5 & $62.3$  &  $56.2$  &  $55.8$  &  $58.0$  &  $53.2$  &  $54.9$  &  $52.5$  &  $56.1$\\

%    \midrule[0.8pt]
% \multirow{2}{*}{$\text{Llama}_{\text{7B}}$}
%             & 0 & $67.8$  &  $60.9$  &  $58.1$  &  $60.0$  &  $55.4$  &  $53.6$  &  $53.4$  &  $58.4$\\
%             & 5 & $55.3$     & $55.3$     & $55.3$     & $55.3$     & $55.3$     & $55.3$     & $55.3$     & $55.3$     \\
%     \cdashlinelr{1-10}
% \multirow{2}{*}{$\text{Llama}_{\text{7B}}$ +AFP}
%             & 0 & $55.3$     & $55.3$     & $55.3$     & $55.3$     & $55.3$     & $55.3$     & $55.3$     & $55.3$     \\
%             & 5 & $55.3$     & $55.3$     & $55.3$     & $55.3$     & $55.3$     & $55.3$     & $55.3$     & $55.3$     \\
    
\bottomrule[1.2pt]
\end{tabu}

\caption{\label{tab:all_pawsx} In-context learning results on PAWS-X across all languages. ``\textbf{High}'' and ``\textbf{Medium}'' denotes the available amount of linguistic resources. ${}^{\dagger}$ denotes the unseen language in the pre-training corpus of BLOOM.
}

\vspace{-2mm}

\end{table*}
\begin{table*}[thp]

\renewcommand\arraystretch{1.3}

\centering
\scriptsize

\setlength{\tabcolsep}{0.8mm}

% \vspace{2mm}

 \begin{tabu}{l|c|cc|ccccc|ccc|cc|c}
 
 \toprule[1.2pt]
  \multicolumn{2}{c}{ } & \multicolumn{2}{c}{\textbf{High}} & \multicolumn{5}{c}{\textbf{Medium}} & \multicolumn{3}{c}{\textbf{Low}} & \multicolumn{2}{c}{\textbf{Ex-Low}} &\\
  \cmidrule(r){3-4} \cmidrule(r){5-9} \cmidrule(r){10-12} \cmidrule(r){13-14} \noalign{\smallskip}
\multicolumn{1}{c}{\textbf{Model}}& \textbf{\#shot} & EN & ZH & ID & IT${}^{\dagger}$  & TH${}^{\dagger}$  & TR${}^{\dagger}$  & VI & ET${}^{\dagger}$  & SW & TA & HT${}^{\dagger}$  & QU${}^{\dagger}$  &\multicolumn{1}{c}{\textbf{Avg}}\\

   \midrule[0.8pt]
\multirow{2}{*}{$\text{XGLM}_{\text{7.5B}}$}
            & 0 & $69.4$  &  $62.4$  &  $63.0$  &  $56.0$  &  $62.0$  &  $56.6$  &  $61.4$  &  $57.4$  &  $58.2$  &  $56.2$  &  $56.6$  &  $48.0$  &  $58.9$\\
            & 5 & $74.6$  &  $63.2$  &  $62.6$  &  $57.6$  &  $62.4$  &  $58.4$  &  $66.2$  &  $58.6$  &  $57.2$  &  $57.2$  &  $54.8$  &  $51.8$  &  $60.4$\\
    \cdashlinelr{1-15}
\multirow{2}{*}{$\text{XGLM}_{\text{7.5B}}$ +AFP}
            & 0 & $71.0$  &  $65.2$  &  $64.2$  &  $58.6$  &  $64.2$  &  $60.2$  &  $63.8$  &  $59.8$  &  $60.4$  &  $58.0$  &  $58.4$  &  $52.2$  &  $\textbf{61.3}$\\
            & 5 & $76.6$  &  $66.2$  &  $63.4$  &  $59.6$  &  $64.2$  &  $61.0$  &  $67.4$  &  $61.2$  &  $60.6$  &  $58.4$  &  $56.2$  &  $53.4$  &  $\textbf{62.4}$\\
    
  \midrule[0.8pt]
\multirow{2}{*}{$\text{BLOOMZ}_{\text{7.1B}}$}
            & 0 & $61.2$  &  $57.6$  &  $59.4$  &  $49.4$  &  $53.2$  &  $55.0$  &  $58.2$  &  $49.2$  &  $53.6$  &  $46.0$  &  $43.4$  &  $51.2$  &  $53.1$\\
            & 5 & $62.4$  &  $59.8$  &  $61.0$  &  $49.4$  &  $51.6$  &  $54.2$  &  $61.8$  &  $47.2$  &  $52.2$  &  $58.4$  &  $48.4$  &  $49.2$  &  $54.6$\\
    \cdashlinelr{1-15}

\multirow{2}{*}{$\text{BLOOM}_{\text{7.1B}}$}
            & 0 & $58.0$  &  $54.2$  &  $59.2$  &  $48.6$  &  $52.6$  &  $53.8$  &  $59.0$  &  $48.0$  &  $53.2$  &  $45.0$  &  $46.0$  &  $49.4$  &  $52.3$\\
            & 5 & $58.4$  &  $54.8$  &  $60.0$  &  $50.2$  &  $52.8$  &  $53.4$  &  $57.8$  &  $47.6$  &  $54.6$  &  $53.8$  &  $46.6$  &  $50.0$  &  $53.3$\\
    \cdashlinelr{1-15}
\multirow{2}{*}{$\text{BLOOM}_{\text{7.1B}}$ +AFP}
            & 0 & $59.4$  &  $55.8$  &  $61.6$  &  $53.2$  &  $54.4$  &  $55.2$  &  $60.6$  &  $51.2$  &  $54.2$  &  $54.4$  &  $48.6$  &  $51.8$  &  $\textbf{55.0}$\\
            & 5 & $61.0$  &  $56.8$  &  $61.0$  &  $51.4$  &  $54.2$  &  $54.6$  &  $59.6$  &  $49.8$  &  $55.2$  &  $57.2$  &  $50.2$  &  $50.4$  &  $\textbf{55.1}$\\

%    \midrule[0.8pt]
% \multirow{2}{*}{$\text{Llama}_{\text{7B}}$}
%             & 0 & $55.3$     & $55.3$     & $55.3$     & $55.3$     & $55.3$     & $55.3$     & $55.3$     & $55.3$     & $55.3$     & $55.3$     & $55.3$     & $55.3$     & $55.3$     \\
%             & 5 & $55.3$     & $55.3$     & $55.3$     & $55.3$     & $55.3$     & $55.3$     & $55.3$     & $55.3$     & $55.3$     & $55.3$     & $55.3$     & $55.3$     & $55.3$     \\
%     \cdashlinelr{1-15}
% \multirow{2}{*}{$\text{Llama}_{\text{7B}}$ +AFP}
%             & 0 & $55.3$     & $55.3$     & $55.3$     & $55.3$     & $55.3$     & $55.3$     & $55.3$     & $55.3$     & $55.3$     & $55.3$     & $55.3$     & $55.3$     & $55.3$     \\
%             & 5 & $55.3$     & $55.3$     & $55.3$     & $55.3$     & $55.3$     & $55.3$     & $55.3$     & $55.3$     & $55.3$     & $55.3$     & $55.3$     & $55.3$     & $55.3$     \\
    
\bottomrule[1.2pt]
\end{tabu}

\caption{\label{tab:all_xcopa} In-context learning results on XCOPA across all languages. ``\textbf{High}'', ``\textbf{Medium}'', ``\textbf{Low}'' and ``\textbf{Ex-Low}'' denotes the available amount of linguistic resources. ${}^{\dagger}$ denotes the unseen language in the pre-training corpus of BLOOM.
}

\vspace{-2mm}

\end{table*}
\begin{table*}[thp]

\renewcommand\arraystretch{1.3}

\centering
\scriptsize

\setlength{\tabcolsep}{0.8mm}

% \vspace{2mm}

 \begin{tabu}{l|c|cccc|cc|ccc|cc|c}
 
 \toprule[1.2pt]
  \multicolumn{2}{c}{ } & \multicolumn{4}{c}{\textbf{High}} & \multicolumn{2}{c}{\textbf{Medium}} & \multicolumn{3}{c}{\textbf{Low}} & \multicolumn{2}{c}{\textbf{Ex-Low}} &\\
  \cmidrule(r){3-6} \cmidrule(r){7-8} \cmidrule(r){9-11} \cmidrule(r){12-13} \noalign{\smallskip}
\multicolumn{1}{c}{\textbf{Model}}& \textbf{\#shot} & EN & ES & RU${}^{\dagger}$  & ZH & AR & ID & HI & SW & TE & EU & MY${}^{\dagger}$  &\multicolumn{1}{c}{\textbf{Avg}}\\

   \midrule[0.8pt]
\multirow{2}{*}{$\text{XGLM}_{\text{7.5B}}$}
            & 0 & $69.2$  &  $64.0$  &  $63.4$  &  $59.5$  &  $56.2$  &  $63.0$  &  $59.0$  &  $59.2$  &  $60.2$  &  $57.4$  &  $55.1$  &  $60.6$\\
            & 5 & $73.7$  &  $63.6$  &  $63.6$  &  $59.2$  &  $54.4$  &  $62.2$  &  $59.4$  &  $58.5$  &  $58.7$  &  $56.9$  &  $55.7$  &  $60.5$\\
    \cdashlinelr{1-14}
\multirow{2}{*}{$\text{XGLM}_{\text{7.5B}}$ +AFP}
            & 0 & $70.7$  &  $65.9$  &  $65.7$  &  $62.5$  &  $58.3$  &  $64.1$  &  $60.1$  &  $60.5$  &  $61.1$  &  $60.0$  &  $57.7$  &  $\textbf{62.4}$\\
            & 5 & $74.7$  &  $67.3$  &  $67.4$  &  $62.8$  &  $58.2$  &  $67.1$  &  $61.4$  &  $61.6$  &  $60.7$  &  $59.8$  &  $57.6$  &  $\textbf{62.5}$\\
    
  \midrule[0.8pt]
\multirow{2}{*}{$\text{BLOOMZ}_{\text{7.1B}}$}
            & 0 & $73.7$  &  $64.6$  &  $52.6$  &  $62.1$  &  $60.3$  &  $62.4$  &  $59.2$  &  $55.3$  &  $57.7$  &  $51.9$  &  $48.4$  &  $58.9$\\
            & 5 & $76.9$  &  $65.4$  &  $53.1$  &  $63.9$  &  $60.9$  &  $67.4$  &  $62.8$  &  $57.3$  &  $59.2$  &  $56.4$  &  $47.5$  &  $61.0$\\
    \cdashlinelr{1-14}

\multirow{2}{*}{$\text{BLOOM}_{\text{7.1B}}$}
            & 0 & $70.4$  &  $59.4$  &  $53.5$  &  $64.3$  &  $59.7$  &  $59.6$  &  $58.8$  &  $52.1$  &  $54.7$  &  $50.7$  &  $47.5$  &  $57.3$\\
            & 5 & $73.5$  &  $64.1$  &  $51.7$  &  $64.8$  &  $60.0$  &  $64.6$  &  $61.3$  &  $53.1$  &  $56.9$  &  $53.7$  &  $47.3$  &  $59.2$\\
    \cdashlinelr{1-14}
\multirow{2}{*}{$\text{BLOOM}_{\text{7.1B}}$ +AFP}
            & 0 & $70.8$  &  $67.6$  &  $54.3$  &  $67.6$  &  $62.3$  &  $65.2$  &  $62.3$  &  $57.6$  &  $57.0$  &  $59.2$  &  $50.0$  &  $\textbf{61.3}$\\
            & 5 & $75.4$  &  $66.4$  &  $54.8$  &  $69.0$  &  $64.0$  &  $70.5$  &  $63.1$  &  $56.9$  &  $58.1$  &  $59.4$  &  $49.8$  &  $\textbf{62.5}$\\

%    \midrule[0.8pt]
% \multirow{2}{*}{$\text{Llama}_{\text{7B}}$}
%             & 0 & $55.3$     & $55.3$     & $55.3$     & $55.3$     & $55.3$     & $55.3$     & $55.3$     & $55.3$     & $55.3$     & $55.3$     & $55.3$     & $55.3$     \\
%             & 5 & $55.3$     & $55.3$     & $55.3$     & $55.3$     & $55.3$     & $55.3$     & $55.3$     & $55.3$     & $55.3$     & $55.3$     & $55.3$     & $55.3$     \\
%     \cdashlinelr{1-14}
% \multirow{2}{*}{$\text{Llama}_{\text{7B}}$ +AFP}
%             & 0 & $55.3$     & $55.3$     & $55.3$     & $55.3$     & $55.3$     & $55.3$     & $55.3$     & $55.3$     & $55.3$     & $55.3$     & $55.3$     & $55.3$     \\
%             & 5 & $55.3$     & $55.3$     & $55.3$     & $55.3$     & $55.3$     & $55.3$     & $55.3$     & $55.3$     & $55.3$     & $55.3$     & $55.3$     & $55.3$     \\
    
\bottomrule[1.2pt]
\end{tabu}

\caption{\label{tab:all_xstorycloze} In-context learning results on XStoryCloze across all languages. ``\textbf{High}'', ``\textbf{Medium}'', ``\textbf{Low}'' and ``\textbf{Ex-Low}'' denotes the available amount of linguistic resources. ${}^{\dagger}$ denotes the unseen language in the pre-training corpus of BLOOM.
}

\vspace{-2mm}

\end{table*}
\begin{table*}[thp]

\renewcommand\arraystretch{1.3}

\centering
\scriptsize

\setlength{\tabcolsep}{0.8mm}

% \vspace{2mm}

 \begin{tabu}{l|c|ccccc|c|c}
 
 \toprule[1.2pt]
  \multicolumn{2}{c}{ } & \multicolumn{5}{c}{\textbf{High}} & \multicolumn{1}{c}{\textbf{Medium}} &\\
  \cmidrule(r){3-7} \cmidrule(r){8-8} \noalign{\smallskip}
\multicolumn{1}{c}{\textbf{Model}}& \textbf{\#shot} & EN & FR & RU${}^{\dagger}$  & ZH & JA${}^{\dagger}$  & PT &\multicolumn{1}{c}{\textbf{Avg}}\\

   \midrule[0.8pt]
\multirow{2}{*}{$\text{XGLM}_{\text{7.5B}}$}
            & 0 & $62.8$  &  $59.0$  &  $58.7$  &  $73.8$  &  $66.4$  &  $62.4$  &  $63.9$\\
            & 5 & $66.4$  &  $62.7$  &  $60.6$  &  $73.2$  &  $62.6$  &  $62.7$  &  $64.7$\\
    \cdashlinelr{1-9}
\multirow{2}{*}{$\text{XGLM}_{\text{7.5B}}$ +AFP}
            & 0 & $64.6$  &  $61.4$  &  $60.3$  &  $75.2$  &  $66.6$  &  $64.6$  &  $\textbf{65.5}$\\
            & 5 & $70.2$  &  $63.9$  &  $63.2$  &  $74.2$  &  $63.9$  &  $64.6$  &  $\textbf{66.7}$\\
  \midrule[0.8pt]
\multirow{2}{*}{$\text{BLOOMZ}_{\text{7.1B}}$}
            & 0 & $64.1$  &  $59.0$  &  $56.5$  &  $66.1$  &  $51.6$  &  $62.7$  &  $60.0$\\
            & 5 & $66.9$  &  $60.2$  &  $54.3$  &  $68.5$  &  $52.6$  &  $60.1$  &  $60.4$\\
    \cdashlinelr{1-9}

\multirow{2}{*}{$\text{BLOOM}_{\text{7.1B}}$}
            & 0 & $60.6$  &  $56.6$  &  $55.2$  &  $71.4$  &  $51.7$  &  $58.6$  &  $59.0$\\
            & 5 & $63.8$  &  $57.8$  &  $55.6$  &  $67.7$  &  $51.8$  &  $58.6$  &  $59.2$\\
            
    \cdashlinelr{1-9}
\multirow{2}{*}{$\text{BLOOM}_{\text{7.1B}}$ +AFP}
            & 0 & $62.1$  &  $57.8$  &  $58.4$  &  $72.2$  &  $53.6$  &  $62.4$  &  $\textbf{61.1}$\\
            & 5 & $64.8$  &  $61.4$  &  $56.2$  &  $68.5$  &  $52.6$  &  $59.7$  &  $\textbf{60.5}$\\

%    \midrule[0.8pt]
% \multirow{2}{*}{$\text{Llama}_{\text{7B}}$}
%             & 0 & $55.3$     & $55.3$     & $55.3$     & $55.3$     & $55.3$     & $55.3$     & $55.3$     \\
%             & 5 & $55.3$     & $55.3$     & $55.3$     & $55.3$     & $55.3$     & $55.3$     & $55.3$     \\
%     \cdashlinelr{1-9}
% \multirow{2}{*}{$\text{Llama}_{\text{7B}}$ +AFP}
%             & 0 & $55.3$     & $55.3$     & $55.3$     & $55.3$     & $55.3$     & $55.3$     & $55.3$     \\
%             & 5 & $55.3$     & $55.3$     & $55.3$     & $55.3$     & $55.3$     & $55.3$     & $55.3$     \\
    
\bottomrule[1.2pt]
\end{tabu}

\caption{\label{tab:all_xwinograd} In-context learning results on XWinograd across all languages. ``\textbf{High}'' and ``\textbf{Medium}'' denotes the available amount of linguistic resources. ${}^{\dagger}$ denotes the unseen language in the pre-training corpus of BLOOM.
}

\vspace{-2mm}

\end{table*}

\section{Task Descriptions and Prompt Templates}
\label{appendix:prompt_templates}
To comprehensively evaluate our models, six datasets across four tasks are adopted in this work. Table \ref{tab:datasets} shows the statistics of all datasets used. 
It is noted that the original COPA dataset \citep{roemmele2011choice} in English is also included in the evaluation. 
Most templates of prompt follow the ones in \citet{lin-etal-2022-shot}. 
\begin{table*}[htp]

\renewcommand\arraystretch{1.3}

\centering
\scriptsize

\setlength{\tabcolsep}{2.3mm}

% \vspace{0.2cm}

 \begin{tabu}{c|ccccrrr}
 
 \toprule[1.2pt]

\multicolumn{1}{c}{\textbf{Task}} & \multicolumn{1}{c}{\textbf{Dataset}}    &\multicolumn{1}{c}{\textbf{\#Lang}} & \multicolumn{1}{c}{\textbf{Data Curation}} & \multicolumn{1}{c}{\textbf{Metric}} & \multicolumn{1}{r}{\textbf{\#Train}}  & \multicolumn{1}{r}{\textbf{\#Dev}}  &\multicolumn{1}{r}{\textbf{\#Test}}   \\

   \midrule[0.8pt]
Natural Language Inference          &$\text{XNLI}$                          & $15$  &$\text{Translation}$   & Accuracy  & $-$           & $2,490$       & $5,010$  \\
\cdashlinelr{1-8}
Paraphrase Detection                &$\text{PAWS-X}$                        & $7$   &$\text{Aligned}$       & Accuracy  & $-$           & $2,000$       & $2,000$ \\
\cdashlinelr{1-8}
\multirow{3}{*}{Reasoning}
                                    &$\text{XCOPA}$                         & $12$  &$\text{Translation}$   & Accuracy  & $33,810$      & $100$         & $500$     \\
                                    &$\text{XStoryCloze}$                   & $11$  &$\text{Translation}$   & Accuracy  & $361$         & $-$           & $1,511$    \\
                                    &$\text{XWinograd}$                     & $6$   &$\text{Translation}$   & Accuracy  & $-$           & $-$           & $2,325^{\ddagger}$    \\
\cdashlinelr{1-8}
% \multirow{2}{*}{Multilingual Summarization}          
                                    % &$\text{En2ZhSum\&Zh2EnSum}$            & $2$   &$\text{Translation}$   & ROUGE & $2,058,400$   & $6,000$       & $6,000$  \\
                                    % &$\text{XLSUM}$                         & $44$  &$\text{Aligned}$       & ROUGE & $1,122,857$   & $114,198$     & $114,198$  \\
% \cdashlinelr{1-8}
Multilingual Machine Translation    &$\text{FLORES-101}$                    & $101$ &$\text{Aligned}$       & BLEU  & $-$           & $997$         & $1,012$      \\
\specialrule{0em}{0pt}{0pt}

\bottomrule[1.2pt]
\end{tabu}

\caption{\label{tab:datasets} Statistic of evaluation datasets used. ${}^{\ddagger}$ denotes the number of English samples, as the number of test samples in XWinograd varies across languages. 
}

\vspace{-2mm}

\end{table*}

\paragraph{Natural Language Inference} This task aims to determine the semantic relationship between the premise and hypothesis.
Table \ref{tab:xnli_examples} illustrates the template and 3-shot example used in our evaluation for this task. 
\begin{table*}[ht]

\renewcommand\arraystretch{1.5}

\centering
\
\setlength{\tabcolsep}{1.2mm}

% \vspace{0.2cm}

 \begin{tabu}{l|l}
 
 \toprule[1.2pt]
% \multicolumn{2}{c}{\textit{Example for English samples}} \\
%  \cdashlinelr{1-2}

\multicolumn{1}{c}{\textbf{Template}} & \multicolumn{1}{c}{\textbf{Candidate Verbalizer}} \\
\midrule[0.8pt]
\{Premise\}\textit{, right?} \{\underline{Label}\}, \{Hypothesis\} & Entailment$\to$Yes, Neural$\to$Also, Contradiction$\to$No \\

\cdashlinelr{1-2}

\multicolumn{2}{c}{\textit{3-shot Example in English}} \\
 
\cdashlinelr{1-2}

\multicolumn{2}{l}{\text{We ask every nation to join us.}\textit{, right?} \underline{Also}\text{, We need at least 10 countries to join us.}\textit{\textless/s\textgreater}} \\
\multicolumn{2}{l}{\text{One of the benefits we get of course is travel.}\textit{, right?} \underline{Yes}\text{, Traveling is one perk we get.}\textit{\textless/s\textgreater}} \\
\multicolumn{2}{l}{\text{Serious crime down, but murders increase.}\textit{, right?} \underline{Yes}\text{, There has been a rise in murders.}\textit{\textless/s\textgreater}} \\
\multicolumn{2}{l}{\text{So I'm not really sure why.}\textit{, right?} {\color{red} No}\text{, I am certain as to the reason why.}} \\

\bottomrule[1.2pt]
\end{tabu}

\caption{\label{tab:xnli_examples} Template and example of 3-shot demonstrations used in the evaluation of XNLI. Connectors are indicated in \textit{italics}. The label for each example is \underline{underlined}. The {\color{red} red} text is the prediction from the model evaluated. 
}

\vspace{-2mm}

\end{table*}

\paragraph{Paraphrase Detection} Models need to evaluate whether the second sentence is a paraphrase of the first sentence in this task. 
The template and 3-shot example adopted are reported in Table \ref{tab:pawsx_examples}.

\begin{table*}[ht]

\renewcommand\arraystretch{1.5}

\centering
\
\setlength{\tabcolsep}{1.2mm}

% \vspace{0.2cm}

 \begin{tabu}{l|l}
 
 \toprule[1.2pt]
% \multicolumn{2}{c}{\textit{Example for English samples}} \\
%  \cdashlinelr{1-2}
 \multicolumn{1}{c}{\textbf{Template}} & \multicolumn{1}{c}{\textbf{Candidate Verbalizer}} \\
\midrule[0.8pt]
\{Sentence 1\}\textit{, right?} \{\underline{Label}\}, \{Sentence 2\} & \multicolumn{1}{c}{True$\to$Yes, False$\to$No}  \\

\cdashlinelr{1-2}

\multicolumn{2}{c}{\textit{3-shot Example in English}} \\
 
\cdashlinelr{1-2}

\multicolumn{2}{l}{\text{Write anywhere , run once}\textit{, right?} \underline{No}\text{, Write anywhere , once run}\textit{\textless/s\textgreater}} \\
\multicolumn{2}{l}{\text{It was Easipower that said :}\textit{, right?}\underline{Yes}\text{, It said that Easipower was ,}\textit{\textless/s\textgreater}} \\
\multicolumn{2}{l}{\text{In 1951 , he died and retired in 1956 .}\textit{, right?} \underline{No}\text{, He died in 1951 and retired in 1956 .}\textit{\textless/s\textgreater}} \\
\multicolumn{2}{l}{\text{Green took over Park 's No .}\textit{, right?} {\color{red} Yes}\text{, Park Green took over No .}} \\

\bottomrule[1.2pt]
\end{tabu}

\caption{\label{tab:pawsx_examples} Examples of 3-shot demonstrations used in the evaluation of PAWS-X. Connectors are indicated in \textit{italics}. The label for each example is \underline{underlined}. The {\color{red} red} text is the prediction from the model evaluated. 
}

% \vspace{-2mm}

\end{table*}

\paragraph{Reasoning} Three popular multilingual reasoning datasets are applied in this task category. 
Given candidate sentences or pronouns mentioned above, models have to select the best one with semantic coherence and comply with the rules of the physics world. 
The detailed templates and examples are presented in Table \ref{tab:xcopa_examples} (XCOPA), Table \ref{tab:xstorycloze_examples} (XStoryCloze) and Table \ref{tab:xwinogrande_examples} (XWinogrande).

\begin{table*}[ht]

\renewcommand\arraystretch{1.5}

\centering
\
\setlength{\tabcolsep}{1.2mm}

% \vspace{0.2cm}

 \begin{tabu}{l|l}
 
 \toprule[1.2pt]
% \multicolumn{2}{c}{\textit{Example for English samples}} \\
%  \cdashlinelr{1-2}

 \multicolumn{1}{c}{\textbf{Template}} & \multicolumn{1}{c}{\textbf{Candidate Verbalizer}} \\
\midrule[0.8pt]
[\textit{cause:$\mid$effect:}] \{Sentence 1\} [\textit{because$\mid$so}] \{\underline{Label}\} & \multicolumn{1}{c}{Identity}  \\

\cdashlinelr{1-2}

\multicolumn{2}{c}{\textit{3-shot Example in English}} \\
 
\cdashlinelr{1-2}
 
\multicolumn{2}{l}{\textit{cause: }\text{The woman resigned.}\textit{because }\underline{She thinks her boss is behaving immorally.}\textit{\textless/s\textgreater}} \\
\multicolumn{2}{l}{\textit{effect: }\text{I pulled the rubber band.}\textit{so }\underline{It stretches out.}\textit{\textless/s\textgreater}} \\
\multicolumn{2}{l}{\textit{cause: }\text{My skin suddenly broke out in a rash.}\textit{because }\underline{I came across poison ivy in my yard.}\textit{\textless/s\textgreater}} \\
\multicolumn{2}{l}{\textit{cause: }\text{The girl pinched her nose.}\textit{because }{\color{red}\text{The baby soiled the diaper.}}} \\

\bottomrule[1.2pt]
\end{tabu}

\caption{\label{tab:xcopa_examples} Examples of 3-shot demonstrations used in the evaluation of XCOPA. Connectors are indicated in \textit{italics}. The label for each example is \underline{underlined}. The {\color{red} red} text is the prediction from the model evaluated. 
}
\vspace{-2mm}

\end{table*}

\begin{table*}[ht]

\renewcommand\arraystretch{1.5}

\centering
\
\setlength{\tabcolsep}{1.2mm}

% \vspace{0.2cm}

 \begin{tabu}{l|l}
 
 \toprule[1.2pt]
% \multicolumn{2}{c}{\textit{Example for English samples}} \\
%  \cdashlinelr{1-2}

 \multicolumn{1}{c}{\textbf{Template}} & \multicolumn{1}{c}{\textbf{Candidate Verbalizer}} \\
\midrule[0.8pt]
\{Sentence 1\} \{Sentence 2\} \{Sentence 3\} \{Sentence 4\} \{\underline{Label}\} & \multicolumn{1}{c}{Identity}  \\

\cdashlinelr{1-2}

\multicolumn{2}{c}{\textit{3-shot Example in English}} \\
 
\cdashlinelr{1-2}
 
\multicolumn{2}{l}{\text{Ava started to notice wrinkles by her eyes. She bought an expensive wrinkle cream. She applied}} \\
\multicolumn{2}{l}{\text{it every night. After a month she checked her eyes out carefully. }\underline{She was happy to see her wrink-}} \\
\multicolumn{2}{l}{\underline{les were gone.}\textit{\textless/s\textgreater}} \\
 % \cdashlinelr{1-2}

\multicolumn{2}{l}{\text{Jenny wanted to learn how to ride a horse. She went to a local horse farm. After a quick lesson,}} \\
\multicolumn{2}{l}{\text{she mounted the horse. A feeling of joy enveloped her as she rode the horse around a ring. }\underline{She}} \\
\multicolumn{2}{l}{\underline{decided to come back soon for another fun lesson.}\textit{\textless/s\textgreater}} \\

\multicolumn{2}{l}{\text{Rick liked eating chocolate oatmeal. But his friend suggested that he use higher quality cocoa }}\\
\multicolumn{2}{l}{\text{powder. Rick was tight about money. But he decided to buy more expensive cocoa powder just}}\\
\multicolumn{2}{l}{\text{once. } \underline{The taste was worth the price.}\textit{\textless/s\textgreater}} \\

\multicolumn{2}{l}{\text{Gordon bought his son a remote control car for Christmas. But he realized that it needed AA}}\\
\multicolumn{2}{l}{\text{batteries. Gordon could not find any. So the next day, he went to the toy store where he bought}}\\
\multicolumn{2}{l}{\text{the car. }{\color{red}\text{He bought a big package of AA batteries.}}} \\

\bottomrule[1.2pt]
\end{tabu}

\caption{\label{tab:xstorycloze_examples} Examples of 3-shot demonstrations used in the evaluation of XStoryCloze. Connectors are indicated in \textit{italics}. The label for each example is \underline{underlined}. The {\color{red} red} text is the prediction from the model evaluated. 
}

\vspace{-2mm}

\end{table*}

\begin{table*}[ht]

\renewcommand\arraystretch{1.5}

\centering
\
\setlength{\tabcolsep}{1.2mm}

% \vspace{0.2cm}

 \begin{tabu}{l|l}
 
 \toprule[1.2pt]
% \multicolumn{2}{c}{\textit{Example for English samples}} \\
%  \cdashlinelr{1-2}

 \multicolumn{1}{c}{\textbf{Template}} & \multicolumn{1}{c}{\textbf{Candidate Verbalizer}} \\
\midrule[0.8pt]
\{Part 1 of Sentence\} \{\underline{Label}\} \{Part 2 of Sentence\} & \multicolumn{1}{c}{Identity}  \\

\cdashlinelr{1-2}

\multicolumn{2}{c}{\textit{3-shot Example in English}} \\
 
\cdashlinelr{1-2}

\multicolumn{2}{l}{\text{Charles Dickinson shot at Andrew Jackson, so }\underline{Charles Dickinson}\text{ started reloading.}\textit{\textless/s\textgreater}} \\
 % \cdashlinelr{1-2}
\multicolumn{2}{l}{\text{The cheetah outran the antelope so }\underline{The cheetah}\text{ got to eat.}\textit{\textless/s\textgreater}} \\
\multicolumn{2}{l}{\text{The lawyer asked the witness a question, but }\underline{The lawyer}\text{ was reluctant to repeat it.}\textit{\textless/s\textgreater}} \\
\multicolumn{2}{l}{\text{The outlet powered the lamp when }{\color{red}{\text{The outlet}}}\text{ had electricity.}} \\

\bottomrule[1.2pt]
\end{tabu}

\caption{\label{tab:xwinogrande_examples} Examples of 3-shot demonstrations used in the evaluation of XWinogrande. Connectors are indicated in \textit{italics}. The label for each example is \underline{underlined}. The {\color{red} red} text is the prediction from the model evaluated. 
}
\vspace{-2mm}

\end{table*}

\paragraph{Multilingual Machine Translation} Given sentences in the source language, models for this task have to generate the corresponding sentences in the target language. 
Table \ref{tab:flores_101} illustrates the template and 3-shot example used in our evaluation for FLORES-101.

\begin{table*}[ht]

\renewcommand\arraystretch{1.5}

\centering
\
\setlength{\tabcolsep}{1.2mm}

% \vspace{0.2cm}

 \begin{tabu}{l|l}
 
 \toprule[1.2pt]
% \multicolumn{2}{c}{\textit{Example for English samples}} \\
%  \cdashlinelr{1-2}
 \multicolumn{1}{c}{\textbf{Template}} & \multicolumn{1}{c}{\textbf{Candidate Verbalizer}} \\
\midrule[0.8pt]
\{Src. Lang.\}\textit{: }\{Src. Sent.\}\textit{ = } \{Tgt. Lang.\}\textit{: } \{\underline{Tgt. Sent.}\}& \multicolumn{1}{c}{Identity}  \\

\cdashlinelr{1-2}

\multicolumn{2}{c}{\textit{3-shot Example in English}} \\
 
\cdashlinelr{1-2}

\multicolumn{2}{l}{\text{English}\textit{: }\text{Since moving to the Catalan-capital, Vidal had played 49 games for the club.}\textit{ = } \text{French}\textit{: }}\\
\multicolumn{2}{l}{\text{Depuis son arrivée dans la capitale catalane, Vidal a joué 49 matchs pour le club.}\textit{\textless/s\textgreater}} \\
\multicolumn{2}{l}{\text{English}\textit{: }\text{Nadal's head to head record against the Canadian is 7–2.}\textit{ = } \text{French}\textit{: } \text{Le score de Nadal}}\\
\multicolumn{2}{l}{\text{en confrontations directes face au Canadien est de 7-2.}\textit{\textless/s\textgreater}} \\
\multicolumn{2}{l}{\text{English}\textit{: }\text{He recently lost against Raonic in the Brisbane Open.}\textit{ = } \text{French}\textit{: }\text{Il a récemment perdu}}\\
\multicolumn{2}{l}{\text{un match contre Raonic durant l'Open de Brisbane.}\textit{\textless/s\textgreater}} \\
\multicolumn{2}{l}{\text{English}\textit{: }\text{Piquet Jr. was sacked after the 2009 Hungarian Grand Prix.}\textit{ = } \text{French}\textit{: }{\color{red} \text{Piquet Jr. a été}}}\\
\multicolumn{2}{l}{\color{red} \text{limogé après le Grand Prix de Hongrie 2009.}} \\

\bottomrule[1.2pt]
\end{tabu}

\caption{\label{tab:flores_101} Examples of 3-shot demonstrations used in the evaluation of FLORES-101. Connectors are indicated in \textit{italics}. The label for each example is \underline{underlined}. The {\color{red} red} text is the prediction from the model evaluated. 
}
\vspace{-2mm}

\end{table*}

\section{Additional Information about Language Code}
\label{appendix:lang_code}
Table \ref{tab:lang_codes} presents more information about the language codes involved in this work. 

\begin{table*}[htp]

% \vspace{0.2cm}

\begin{minipage}[t]{0.49\linewidth}
       \setlength{\tabcolsep}{2mm}
	\centering
	\scriptsize
	\renewcommand\arraystretch{1.25}
	\begin{center}
		% \caption{Ablation study of different training methods on 5 datasets for $\text{XGLM}_{\text{564M}}$.}
		\begin{tabular}{ccc}
			\toprule[1.2pt]  
               \multicolumn{1}{c}{\textbf{ISO 639-1}} & \multicolumn{1}{c}{\textbf{Language}}    &\multicolumn{1}{c}{\textbf{Family}} \\
                \midrule[0.8pt]
			AF      &Afrikaans      &Indo-European\\
                AR      &Arabic      &Afro-Asiatic\\
                AZ      &Azerbaijani      &Turkic\\
                BG${}^{\dagger}$      &Bulgarian      &Indo-European\\
                BN      &Bengali      &Indo-European\\
                CS      &Czech      &Indo-European\\
                DE      &German      &Indo-European\\
                EL${}^{\dagger}$      &Greek, Modern      &Indo-European\\
                EN${}^{\star}$      &English      &Indo-European\\
                ES      &Spanish      &Indo-European\\
                ET      &Estonian      &Uralic\\
                EU${}^{\dagger}$      &Basque      &Language Isolate\\
                FA      &Persian      &Indo-European\\
                FI      &Finnish      &Uralic\\
                FR      &French      &Indo-European\\
                GL      &Galician      &Indo-European\\
                GU      &Gujarati      &Indo-European\\
                HE      &Hebrew      &Afro-Asiatic\\
                HI      &Hindi      &Indo-European\\
                HR      &Croatian      &Indo-European\\
                HT${}^{\dagger}$      &Haitian Creole      &French Creole\\
                ID      &Indonesian      &Austronesian\\
                IT      &Italian      &Indo-European\\
                JA      &Japanese      &Japonic\\
                KA      &Georgian      &Kartvelian\\
                KK      &Kazakh      &Turkic\\
                KM      &Khmer      &Austroasiatic\\
                KO      &Korean      &Koreanic\\
			\bottomrule[1.2pt]
		\end{tabular}
	\end{center}
	% \vspace{-5mm}
\end{minipage}
\begin{minipage}[t]{0.49\linewidth}
       \setlength{\tabcolsep}{2.5mm}
	\centering
	\scriptsize
	%\vspace{-0.4cm}
	\renewcommand\arraystretch{1.2}
	\begin{center}
		% \caption{Ablation study of different training methods on 5 datasets for $\text{BLOOM}_{\text{560M}}$.}
		\begin{tabular}{ccc}
			\toprule[1.2pt]  
               \multicolumn{1}{c}{\textbf{ISO 639-1}} & \multicolumn{1}{c}{\textbf{Language}}    &\multicolumn{1}{c}{\textbf{Family}} \\
                \midrule[0.8pt]
			LT      &Lithuanian      &Indo-European\\
                LV      &Latvian      &Indo-European\\
                MK      &Macedonian      &Indo-European\\
                ML      &Malayalam      &Dravidian\\
                MN      &Mongolian      &Mongolic\\
                MR      &Marathi      &Indo-European\\
                MY      &Burmese      &Sino-Tibetan\\
                NE      &Nepali      &Indo-European\\
                NL      &Dutch      &Indo-European\\
                PL      &Polish      &Indo-European\\
                PS      &Pashto      &Indo-European\\
                PT      &Portuguese      &Indo-European\\
                QU${}^{\dagger}$      &Quechua      &-\\
                RO      &Romanian      &Indo-European\\
                RU      &Russian      &Indo-European\\
                SI      &Sinhala      &Indo-European\\
                SL      &Slovene      &Indo-European\\
                SV      &Swedish      &Indo-European\\
                SW${}^{\star}$      &Swahili      &Niger-Congo\\
                TA      &Tamil      &Dravidian\\
                TE      &Telugu      &Dravidian\\
                TH${}^{\star}$      &Thai      &Kra-Dai\\
                TL      &Tagalog      &Austronesian\\
                TR${}^{\star}$      &Turkish      &Turkic\\
                UK      &Ukrainian      &Indo-European\\
                UR      &Urdu      &Indo-European\\
                VI      &Vietnamese      &Austroasiatic\\
                XH      &Xhosa      &Niger-Congo\\
                ZH${}^{\star}$      &Chinese      &Sino-Tibetan\\
			\bottomrule[1.2pt]
		\end{tabular}
	\end{center}
	% \vspace{-5mm}
\end{minipage}

\caption{\label{tab:lang_codes} Details of Language codes in this work. ${}^{\star}$ denotes the language used in bilingual and 5-language experiments. ${}^{\dagger}$ indicates the languages involved in the multilingual evaluation datasets but not in Bactrian-X. 
}
% \vspace{-2mm}

\end{table*}

\end{document}